\documentclass[runningheads]{llncs}

\usepackage{eccv}

\usepackage{eccvabbrv}

\usepackage{graphicx}
\usepackage{booktabs}
\usepackage{multirow}

\usepackage[accsupp]{axessibility}  

\usepackage{hyperref}

\usepackage{orcidlink}

\def\paperTitle{
A Task is Worth One Word: \\
Learning with Task Prompts for
High-Quality Versatile Image Inpainting}

\def\modelname{PowerPaint}
\def\pinsert{$\mathbf{P_{obj}}$}
\def\premove{$\mathbf{P_{ctxt}}$}
\def\pshape{$\mathbf{P_{shape}}$}

\newif\ifreview 
\newif\ifarxiv 
\newif\ifcamera 
\newif\ifrebuttal

\begin{document}

\title{\paperTitle}

\newcommand{\figteaser}{
\begin{center}
    \centering
    \includegraphics[width=1.0\textwidth]{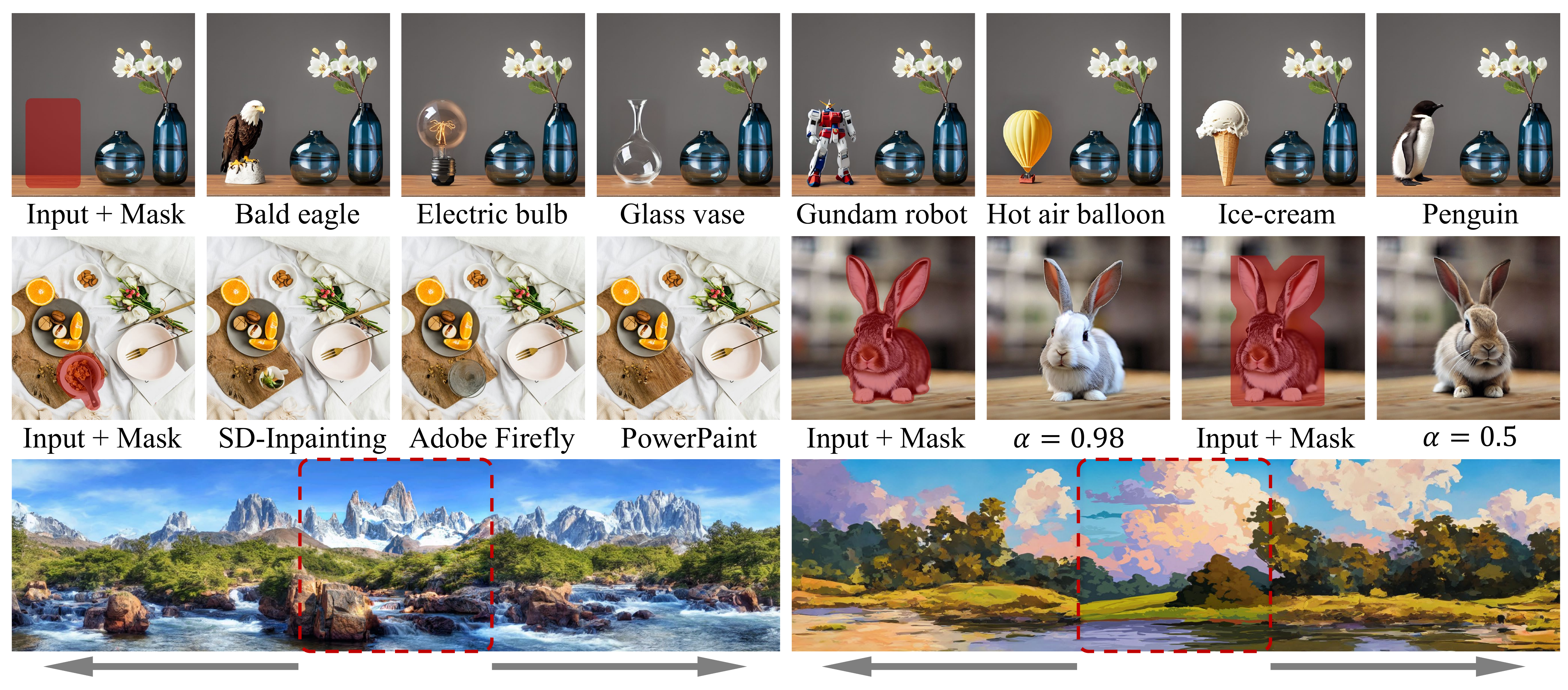}
    \captionof{figure}{
    \textbf{\modelname}
    is the first versatile image inpainting model that simultaneously achieves state-of-the-art results in various inpainting tasks, including text-guided object inpainting, object removal, shape-guided object inpainting with controllable shape-fitting, outpainting, \etc. [Best viewed in color with zoom-in] }
\end{center}%
}

\newcommand{\figoverview}{
\begin{figure*}[t]
    \centering
    \includegraphics[width=1.0\textwidth]{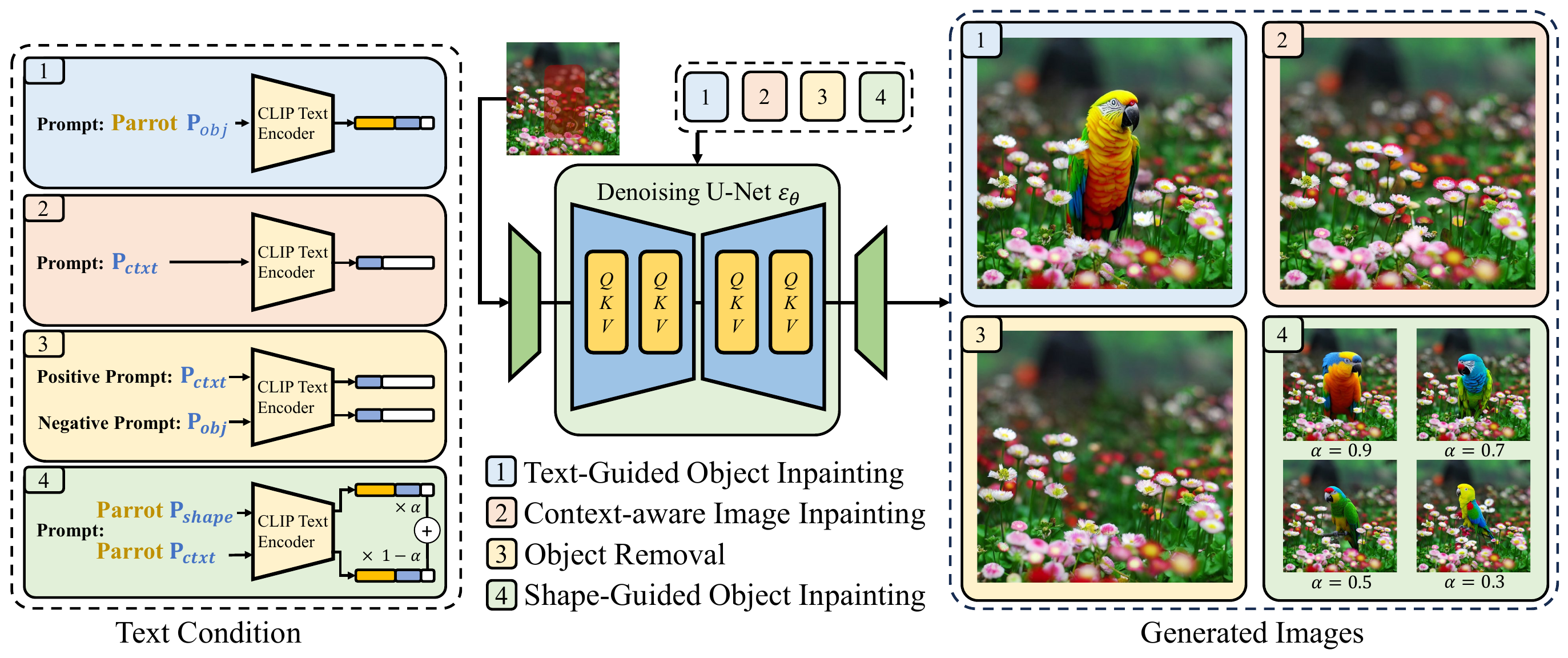}
    \caption{
    \textbf{Overview}. 
    \modelname\ fine-tunes a text-to-image model with two learnable task prompts, i.e., \pinsert\ and \premove, for text-guided object inpainting and context-aware image inpainting, respectively. After training, \pinsert\ can be further used as a negative prompt with classifier-free guidance sampling for effective object removal. We further introduce \pshape\ for shape-guided object inpainting, which can be extended by prompt interpolation with \premove\ to control the degree of shape-fitting for object inpainting.}
    \label{fig:FlowChat}
\end{figure*}}

\newcommand{\figobjremove}{
\begin{figure}[t]
	\small
	\setlength{\tabcolsep}{1.2pt}
	\begin{center}
\resizebox{0.95\textwidth}{!}{
		\begin{tabular}{c c c}
				\includegraphics[width=0.33\linewidth]{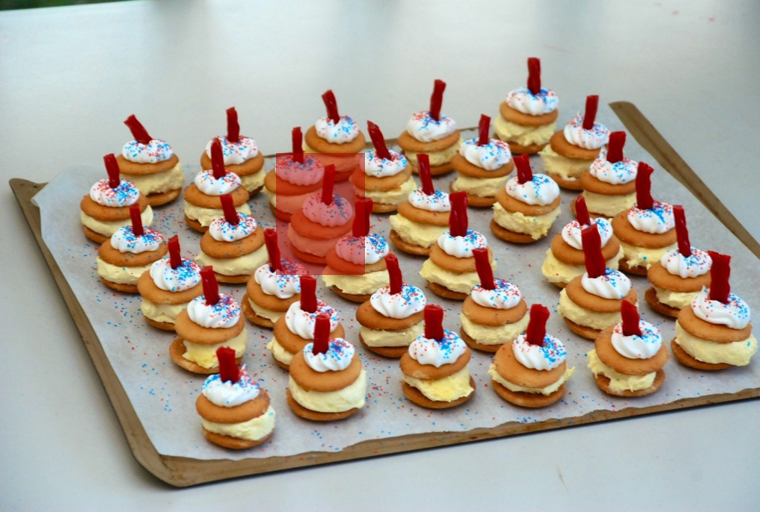} &
				\includegraphics[width=0.33\linewidth]{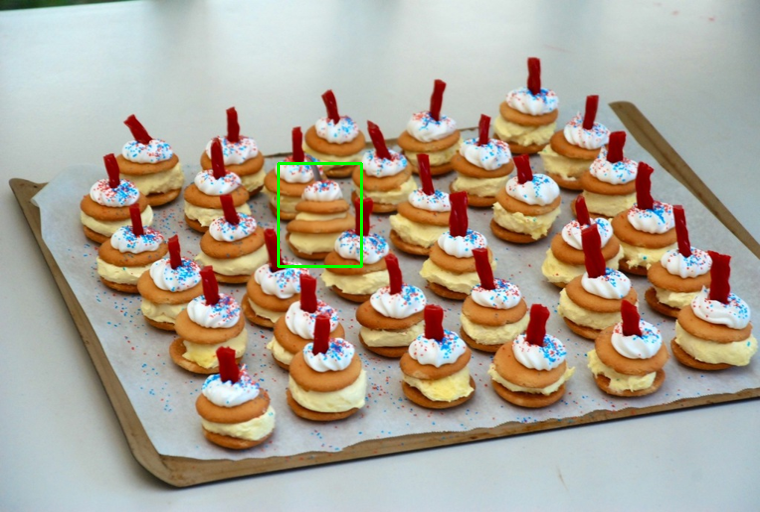} & 
				\includegraphics[width=0.33\linewidth]{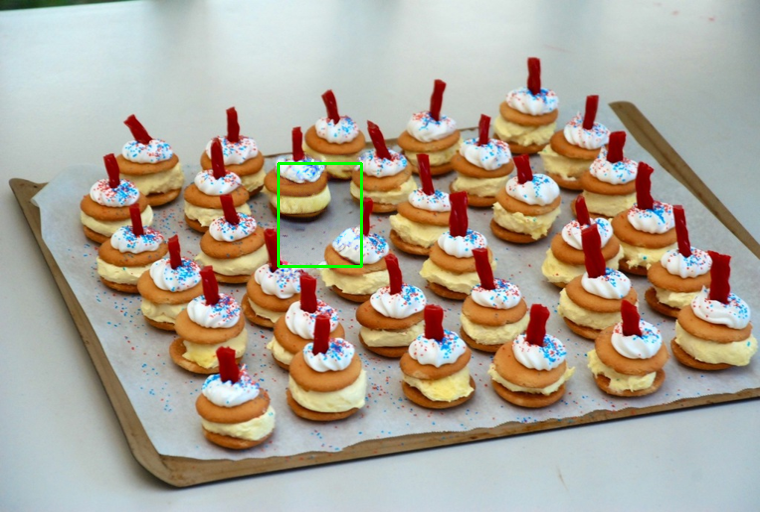}\\
				(a) Original Image & (b) Adobe Firefly & (c) \modelname
		\end{tabular}
  }
	\end{center}
	\caption{To remove objects from crowded image context, the commercial product, Adobe Firefly \cite{adobefirefly2023}, tends to copy from the context (as circled in the green bounding box), while \modelname\ successfully erases the objects.}
	\label{Object erasure}
\end{figure}
}

\newcommand{\figshapeguided}{
\begin{figure}[t]
    \centering
    \includegraphics[width=0.68\textwidth]{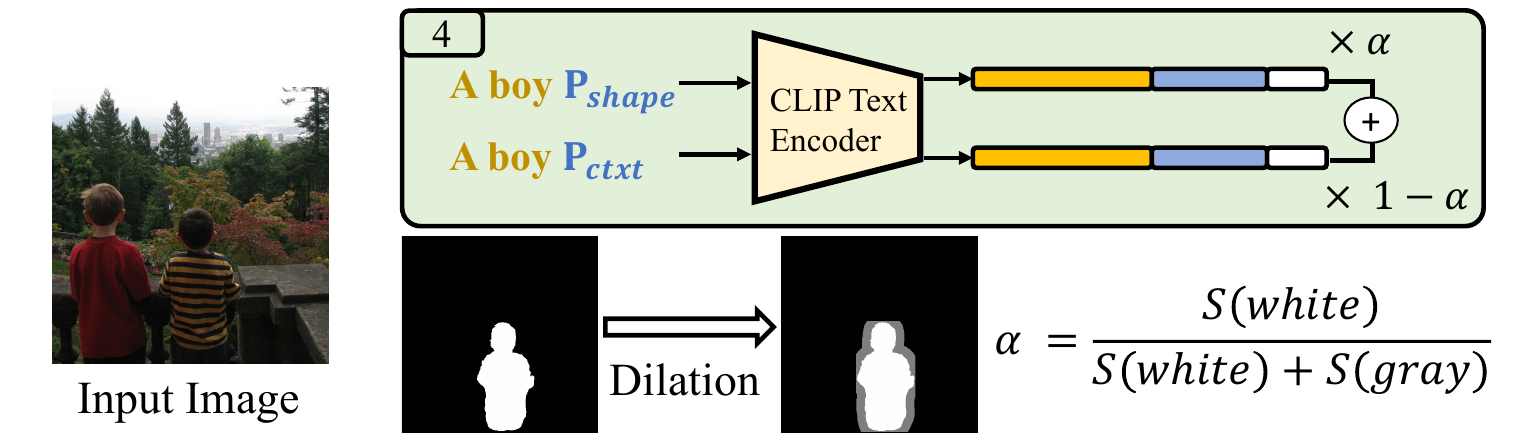}
    \caption{
    \textbf{Illustration of prompt interpolation.} To enable object inpainting with a controllable shape-fitting degree, we randomly expand the object segmentation mask and interpolate \premove\ and \pshape\ according to the expanded area ratio.
    }
    \label{fig:shape_insert}
\end{figure}
}

\newcommand{\tabbbox}{
\begin{table*}[t]
    \centering
  \caption{
  Quantitative comparisons with state-of-the-art models for text-guided object inpainting with bounding box masks. 
  }
  \label{tab:bbox}
\resizebox{0.9\textwidth}{!}{
      \begin{tabular}{c c c c c c c}
    \toprule[1.5pt]
     &   \multicolumn{3}{c}{OpenImages \cite{Kuznetsova_2020}}   & \multicolumn{3}{c}{MSCOCO \cite{lin2014microsoft}}  \\
     &  Local-FID$\downarrow$ & FID$\downarrow$  & CLIP Score$\uparrow$ &Local-FID$\downarrow$ & FID$\downarrow$ & CLIP Score$\uparrow$\\ \hline
    Blended Diffusion \cite{avrahami2022blended} &  35.66 & 8.07 & 27.05 &35.11 &8.66&24.91\\ \hline
    Stable Diffusion \cite{rombach2022high} & 16.75 & 6.59 & 26.79 &20.28 &7.15 &25.13\\ \hline
    ControlNet-Inpainting \cite{zhang2023adding}&  14.74 & 5.66 & 26.87 &17.10 &6.18 &25.01\\ \hline
    SD-Inpainting \cite{rombach2022high} &  12.71 &  4.76 & 26.66 &15.77 &5.65 &24.81\\ \hline
    Smartbrush \cite{xie2023smartbrush} &  9.57 &  4.40 & \textbf{27.61} &11.77 &5.20 &25.89\\
    \hline
    \modelname &  \textbf{9.41} & \textbf{4.38} & {27.56} &\textbf{11.61} &\textbf{5.12} &\textbf{25.95}\\ 
    \bottomrule[1.5pt]
  \end{tabular}
}
\end{table*}
}

\newcommand{\tabobj}{
\begin{table*}[t]
    \centering
  \caption{
  Quantitative comparisons with state-of-the-art models for shape-guided object inpainting with object layout masks. 
  }
  \label{tab:objmask}
\resizebox{0.9\textwidth}{!}{
      \begin{tabular}{c c c c c c c}
    \toprule[1.5pt]
     &   \multicolumn{3}{c}{OpenImages \cite{Kuznetsova_2020}}   & \multicolumn{3}{c}{MSCOCO \cite{lin2014microsoft}}\\
     &  Local-FID$\downarrow$ & FID$\downarrow$  & CLIP Score$\uparrow$ &Local-FID$\downarrow$ & FID$\downarrow$ & CLIP Score$\uparrow$\\ \hline
    Blended Diffusion \cite{avrahami2022blended} &  22.69 & 6.87 &26.60 &26.22 &6.31&24.89\\ \hline
    Stable Diffusion \cite{rombach2022high}&  13.94 & 4.98 & 26.47 &18.23 &5.44 &24.60\\ \hline
    ControlNet-Inpainting \cite{zhang2023adding}&  11.96 & 4.54 & 26.45 &17.15 &5.25 &24.58\\ \hline
    SD-Inpainting \cite{rombach2022high}&  10.73 & 3.93 & 26.31 &16.15 &5.09 &24.53\\ \hline
    Smartbrush \cite{xie2023smartbrush} & 8.01 & 3.65 & 27.10 &11.11&4.67&{25.60}\\ 
    \hline
    \modelname &  \textbf{7.96} & \textbf{3.61} & \textbf{27.14} &\textbf{11.04} &\textbf{4.61} &\textbf{25.62}\\ 
    \bottomrule[1.5pt]
  \end{tabular}
}
\end{table*}
}

\newcommand{\tabinpaint}{
\begin{table}[t]
    \centering
  \caption{
  Quantitative comparisons for context-aware image inpainting on Places2 \cite{zhou2017places}.}
  \label{tab:bgfill}
\hspace{-0.3cm}\resizebox{0.6\textwidth}{!}{
     \begin{tabular}{c c c c c}
    \toprule[1.5pt]
     &   \multicolumn{2}{c}{40-50\% masked}   & \multicolumn{2}{c}{All samples}\\
     & FID $\downarrow$ & LPIPS$\downarrow$ &FID $\downarrow$ & LPIPS$\downarrow$ \\ \hline
    LaMa \cite{suvorov2022lama} &  21.07 & \textbf{0.2133} & 3.48 &\textbf{0.1193}\\ \hline
    LDM-Inpaint \cite{rombach2022high} &  21.42 & 0.2317 & 3.42 &0.1325\\ \hline
    SD-Inpainting \cite{rombach2022high} &  19.73 & 0.2322 & 3.03 &0.1312\\ \hline
    SD-Inpainting(`background') &  19.21 & 0.2290 & 2.82 &0.1293\\ \hline
    SD-Inpainting(`scenery') &  18.93 & 0.2312 & 2.84 &0.1306\\ \hline 
    SmartBrush\cite{xie2023smartbrush}(`scenery') &  87.21 & 0.2812 & 15.21 &0.1579\\ \hline 
    \modelname\ &  \textbf{17.91} & 0.2225 & \textbf{2.59} &0.1263\\ 
    \bottomrule[1.5pt]
  \end{tabular}
}
\end{table}
}

\newcommand{\taboutpaint}{
\begin{table}[t]
    \centering
  \caption{
  Quantitative comparisons for outpainting on Flickr-Scenery \cite{cheng2022inout}.}
  \label{tab:outpaint}
\resizebox{0.55\textwidth}{!}{
     \begin{tabular}{c c c}
    \toprule[1.5pt]
     &  FID $\downarrow$ & Aesthetic Score$\uparrow$ \\ \hline
    LaMa \cite{suvorov2022lama} &  16.63 & 5.01 \\ \hline
    LDM-Inpainting \cite{rombach2022high} & 11.00& 5.10\\\hline 
    SD-Inpainting \cite{rombach2022high} &  58.38 & 5.22 \\ \hline
    SD-Inpainting(`background') &  24.67 & 5.25 \\ \hline
    SD-Inpainting(`scenery') &  13.31 & 5.30 \\ \hline 
    SmartBrush\cite{xie2023smartbrush}(`scenery') &  105.99 & 4.79 \\ \hline 
    \modelname&  \textbf{10.16} & \textbf{5.33} \\ 
    \bottomrule[1.5pt]
  \end{tabular}
    }
\end{table}
}

\newcommand{\tabuser}{
\begin{table}[t]
\centering
\caption{
\textbf{User study.} \modelname\ is preferred by users across three groups of user study, namely object inpainting, object removal and image outpainting.
}
\label{tab:humaneval}
\resizebox{0.75\textwidth}{!}{
\begin{tabular}{c|c|c|c|c}
\toprule[1.0pt]
\multirow{2}{*}{\textbf{Tasks}} & \multirow{2}{*}{Baselines} & \multicolumn{3}{c}{Preference} \\ \cline{3-5}
&&Shape&Text Alignment&Realism\\ \hline
\multirow{3}{*}{Object Inpainting} & PowerPaint & \textbf{48.3\%}&\textbf{50.8\%}&\textbf{40.5\%}\\
&SmartBrush &40.4\% &37.3\%&39.1\%\\
&SD-Inpainting &11.3\%&11.9\%&20.4\%\\
\midrule
\multirow{2}{*}{Object Removal} & Baselines & \modelname & SD-Inpainting & LaMa \\\cline{2-5}
& Preference &\textbf{73.2\%} &11.6\% &15.2\%\\
\midrule
\multirow{2}{*}{Outpainting} & Baselines & \modelname & SD-Inpainting & LDM-Inpainting \\\cline{2-5}
& Preference &\textbf{62.6\%} &22.8\% &14.6\%\\
 \bottomrule[1.0pt]
\end{tabular}}
\end{table}
}

\newcommand{\figbigres}{
\begin{figure*}[!ht]
    \centering
    \includegraphics[width=0.98\textwidth]{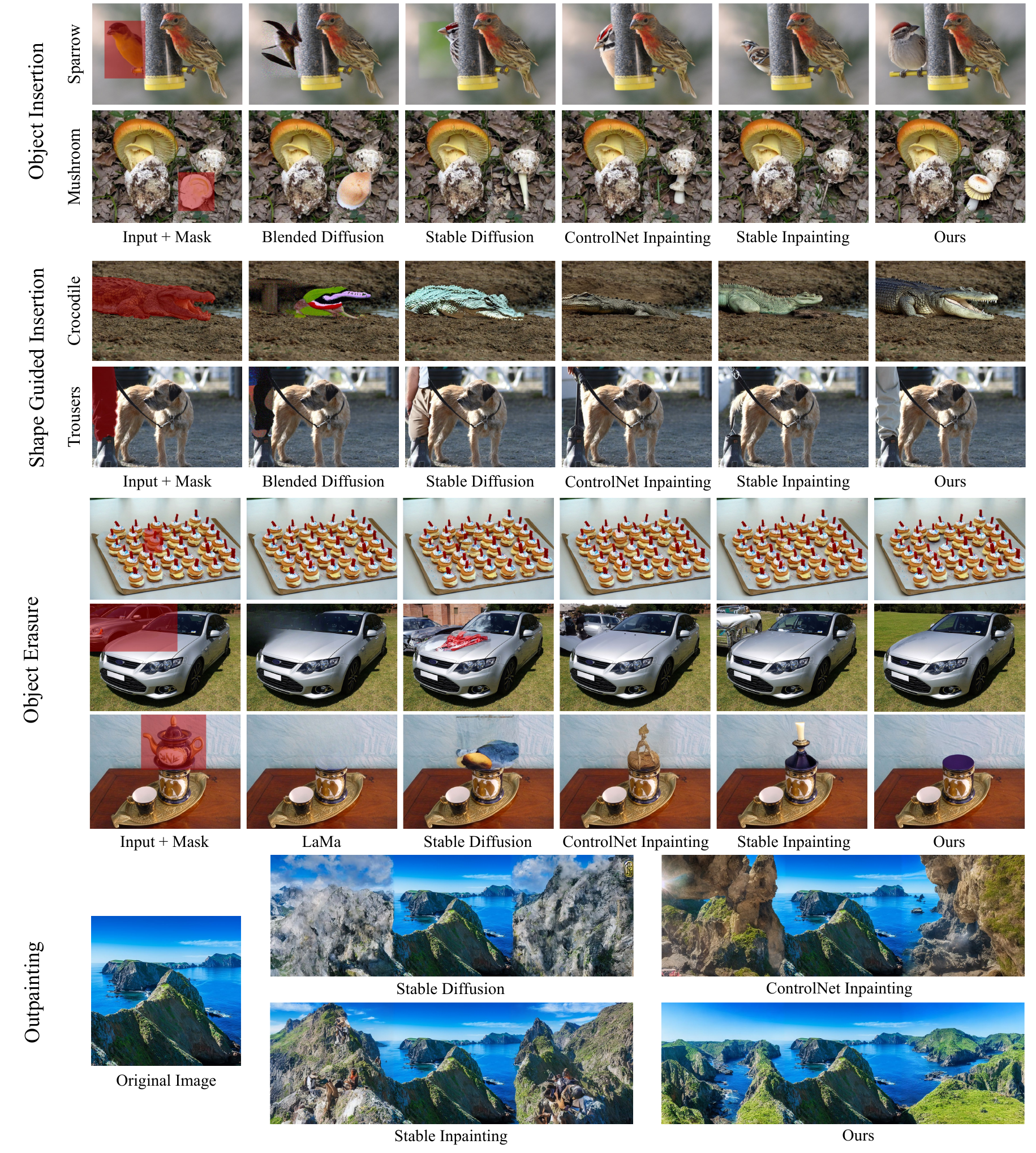}
    \caption{Qualitative comparison with state-of-the-art approaches.}
    \label{fig:exp}
\end{figure*}
}

\newcommand{\figinsert}{
\begin{figure*}[!t]
    \centering
    \hspace{-0.3cm}
    \includegraphics[width=1.018\textwidth]{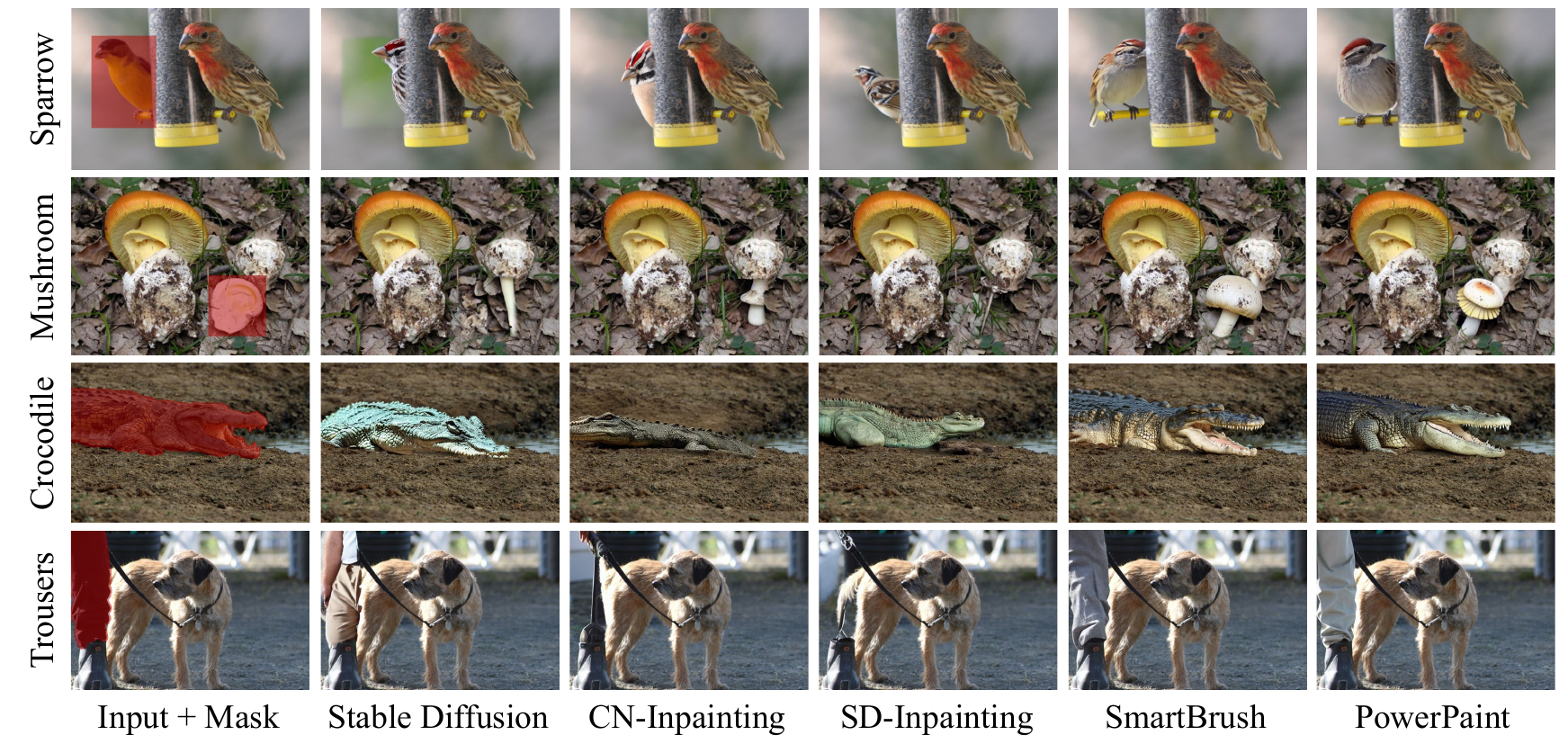}
    \caption{Compared with SOTA approaches, \modelname\ shows better text alignment and visual quality for text-guided object inpainting.}
    \label{fig:biginsert}
\end{figure*}
}

\newcommand{\figinpaint}{
\begin{figure*}[!t]
    \centering
    \includegraphics[width=0.98\textwidth]{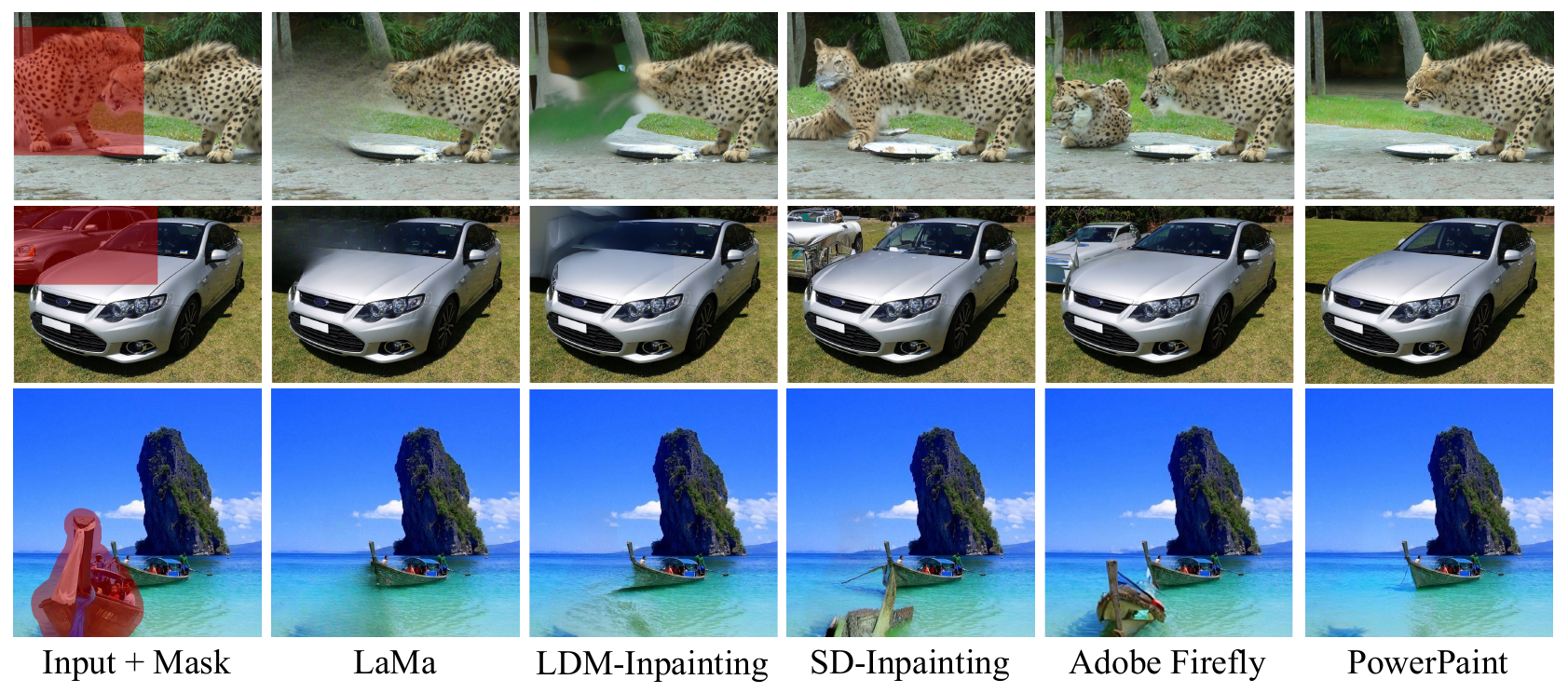}
    \caption{Compared with SOTA approaches, \modelname\ shows better context alignment for context-aware image inpainting.}
    \label{fig:bigcontext}
\end{figure*}
}

\newcommand{\figoutpaint}{
\begin{figure*}[!t]
    \centering
    \includegraphics[width=0.98\textwidth]{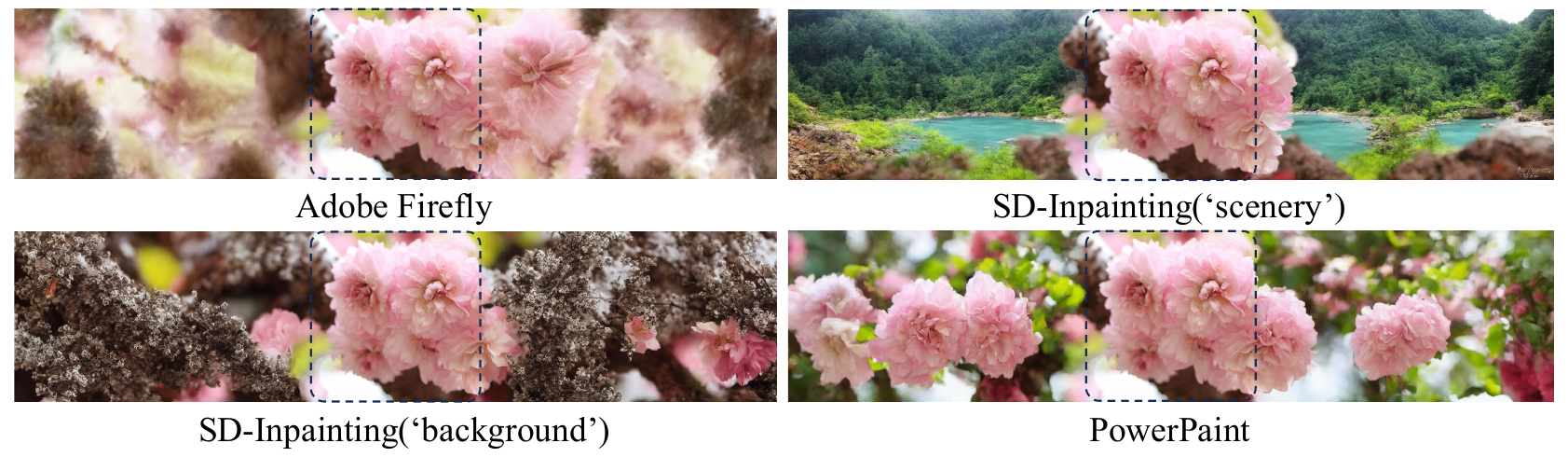}
    \caption{Compared with SOTA approaches, \modelname\ shows more pleasing results for image outpainting with a large expand.}
    \label{fig:bigoutpaint}
\end{figure*}
}

\newcommand{\figuser}{
\begin{figure}
    \centering
    \includegraphics[width=0.9\linewidth]{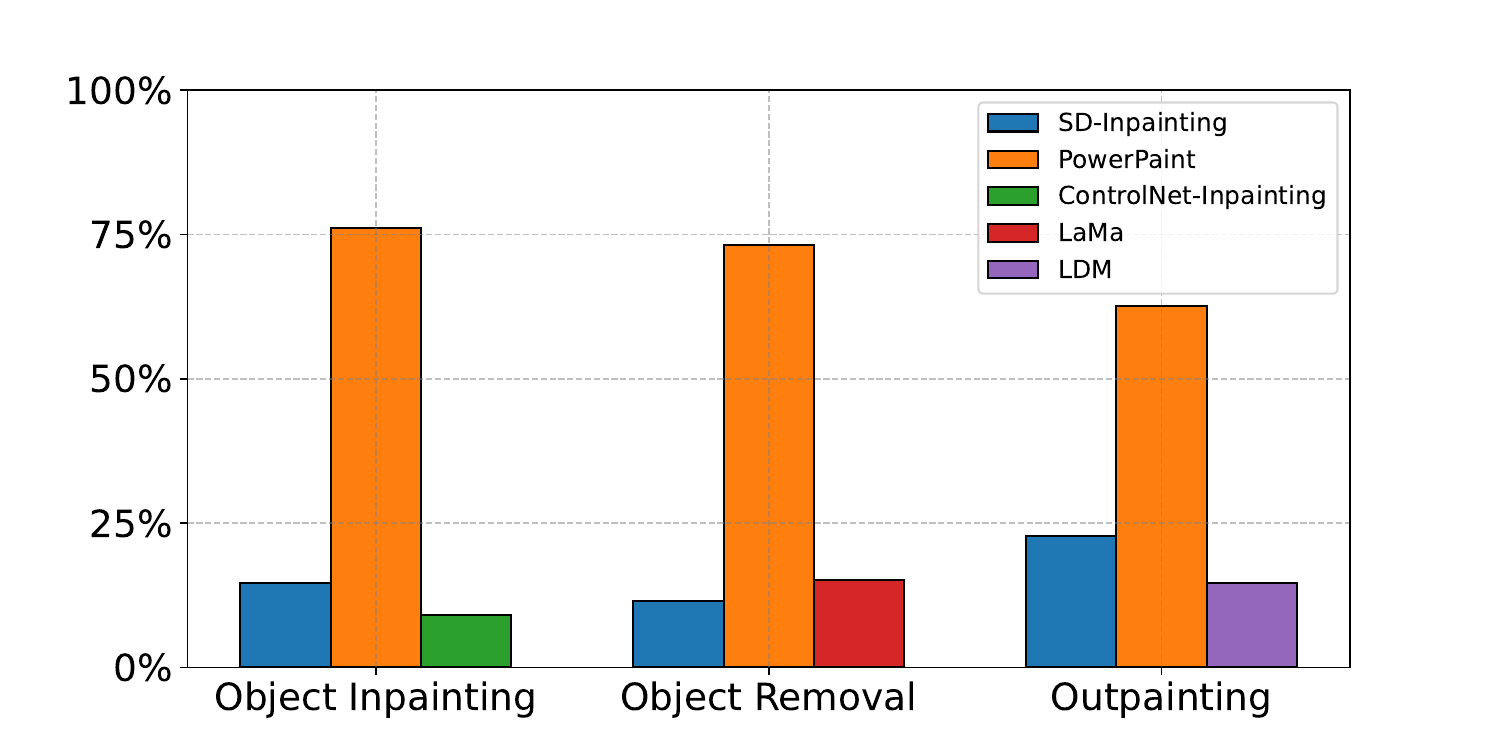}
    \caption{\modelname\ wins the first place in the user studies.}
    \label{fig:userstudy}
\end{figure}
}

\newcommand{\figremove}{
\begin{figure}[t]
    \centering
    \includegraphics[width=0.65\linewidth]{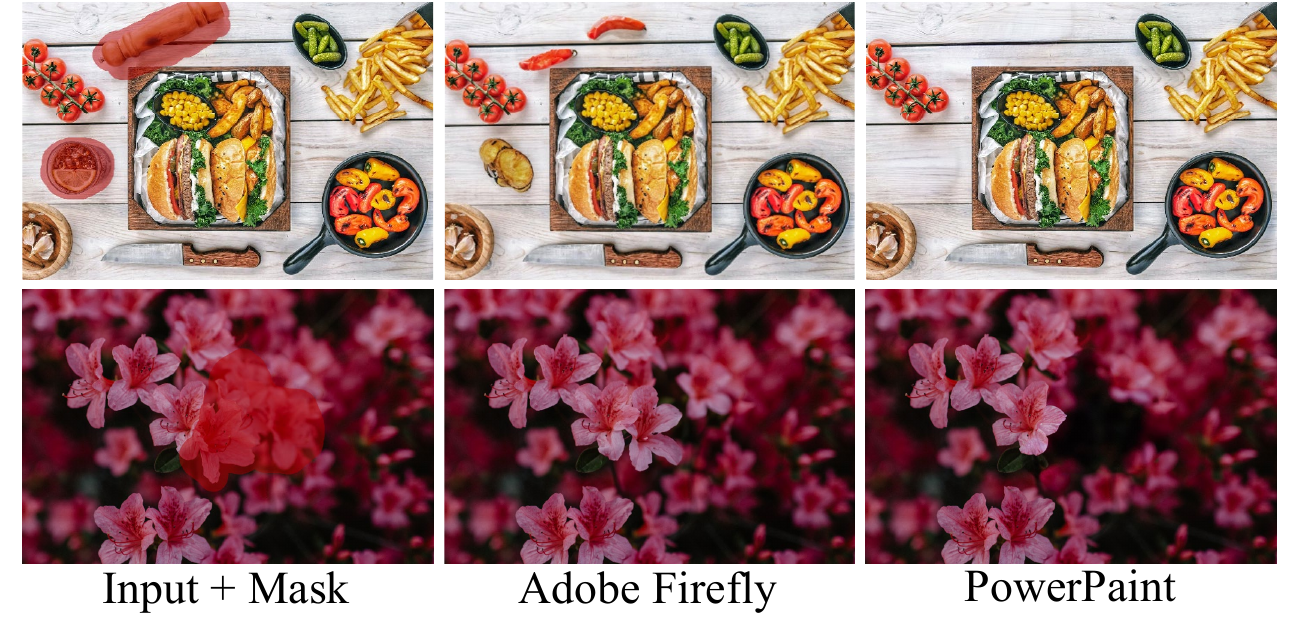}
    \caption{Object removal in comparison with Adobe Firefly \cite{adobefirefly2023}.}
    \label{fig:obj_remove}
\end{figure}
}

\newcommand{\figshape}{
\begin{figure}[t]
    \centering
    \hspace{-0.3cm}\includegraphics[width=0.7\linewidth]{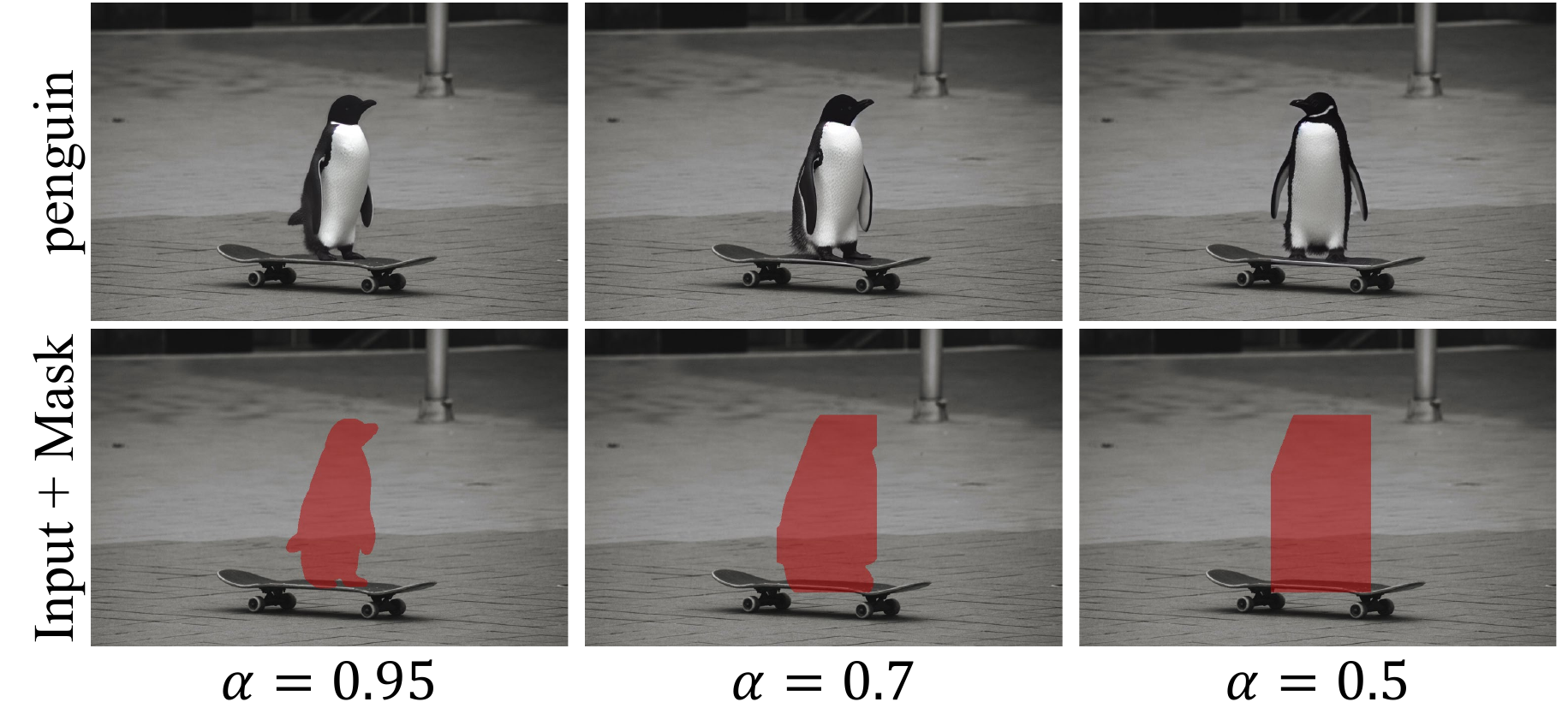}
    \caption{Application of shape-guided object inpainting.}
    \label{fig:shape_guided}
\end{figure}
}

\newcommand{\tababguide}{
\begin{table}[t]
    \centering
  \caption{Ablation study on learnable task prompt. Under the same training strategies, learning with learnable task prompt outperforms the one with unlearnable identifiers.}
  \label{tab:ablation_guide}
\resizebox{0.65\textwidth}{!}{
     \begin{tabular}{c c c c | c c}
    \toprule[1.5pt]
    &\multicolumn{3}{c|}{object inpainting}& \multicolumn{2}{c}{context inpainting}\\
     &Local-FID$\downarrow$ &  FID$\downarrow$ & CLIP Score$\uparrow$ & FID$\downarrow$ & LPIPS$\downarrow$\\ \hline
    identifier & 13.03 & 5.42& 25.72& 3.11 &{0.1312}\\\hline 
    \modelname\ &  \textbf{11.61} & \textbf{5.12}  & \textbf{25.95}&\textbf{2.59} &\textbf{0.1263} \\ 
    \bottomrule[1.5pt]
  \end{tabular}
    }
\end{table}
}

\newcommand{\tabSDinptask}{
\begin{table}[t]
\centering
\caption{Ablation study on single unified model VS task-specific models.
We also include the results of fine-tuning SD-Inpainting on the same fine-tuning dataset.
}
\label{tab:task-specific}
\resizebox{0.85\linewidth}{!}{%
\begin{tabular}{cccc}
\toprule
\textbf{OpenImages [15] / MSCOCO [18]} & \textbf{Local-FID $\downarrow$} & \textbf{FID $\downarrow$}& \textbf{CLIP Score $\uparrow$}\\
\midrule
SD-Inpainting-tuned (bbox) & 12.83 / 15.75 & 4.74 / 5.67& 26.71 / 24.84 \\
Task-specific Model (bbox) & 9.51 / {11.66} & \textbf{4.36} / \textbf{5.11} & \textbf{27.58} / 25.90 \\
PowerPaint (bbox) & \textbf{9.41} / \textbf{11.61} & 4.38 / 5.12 & 27.56 / \textbf{25.95}\\
\hline
SD-Inpainting-tuned (layout) & 10.94 / 16.20 &3.91 / 5.06 & 26.47 / 24.55 \\
Task-specific Model (layout) & 8.03 / \textbf{11.01} & \textbf{3.60} / \textbf{4.59} & \textbf{27.18} / 25.57 \\
PowerPaint (layout) & \textbf{7.96} / 11.04 & 3.61 / 4.61& 27.14 / \textbf{25.62}\\
\bottomrule
\end{tabular}
}
\resizebox{0.8\linewidth}{!}{%
\begin{tabular}{ccc|cc}
\toprule
\textbf{Inpainting / Outpainting} & \textbf{FID $\downarrow$} & \textbf{LPIPS $\downarrow$}& \textbf{FID $\downarrow$} & \textbf{Aesthetic Score $\uparrow$}\\
\midrule
SD-Inpainting-tuned & 2.80 & 0.1302 & 12.65& 5.31 \\
Task-specific Model & 2.68 & 0.1264 & 10.24& 5.31 \\
PowerPaint (ours) & \textbf{2.59} & \textbf{0.1263} & \textbf{10.16} & \textbf{5.33}\\ 
\bottomrule
\end{tabular}
}
  \label{sdinptask}
\end{table}
}
\titlerunning{PowerPaint for High-Quality Versatile Image Inpainting}

\author{Junhao Zhuang\inst{1}* 
Yanhong Zeng\inst{2} 
Wenran Liu\inst{2} 
Chun Yuan\inst{1}$^\dag$
Kai Chen\inst{2}$^\dag$}
\authorrunning{J. Zhuang et al.}

\institute{Tsinghua Shenzhen International Graduate School, Tsinghua University \and
Shanghai Artificial Intelligence Laboratory \\
\email{ \{zhuangjh23@mails, yuanc@sz\}.tsinghua.edu.cn} \\
    \email{ \{zengyanhong, liuwenran, chenkai\}@pjlab.org.cn}}
\maketitle
\figteaser
\let\thefootnote\relax\footnotetext{
* Work done during an internship in Shanghai Artificial Intelligence Lab. \\
$^\dag$ Corresponding authors.}
\begin{abstract}
Advancing image inpainting is challenging as it requires filling user-specified regions for various intents, such as background filling and object synthesis. Existing approaches focus on either context-aware filling or object synthesis using text descriptions. However, achieving both tasks simultaneously is challenging due to differing training strategies.
To overcome this challenge, we introduce \textbf{\modelname}, the first high-quality and versatile inpainting model that excels in multiple inpainting tasks.
First, we introduce learnable task prompts along with tailored fine-tuning strategies to guide the model's focus on different inpainting targets explicitly. This enables \modelname\ to accomplish various inpainting tasks by utilizing different task prompts, resulting in state-of-the-art performance. 
Second, we demonstrate the versatility of the task prompt in \modelname\ by showcasing its effectiveness as a negative prompt for object removal. 
Moreover, we leverage prompt interpolation techniques to enable controllable shape-guided object inpainting, enhancing the model's applicability in shape-guided applications.
Finally, we conduct extensive experiments and applications to verify the effectiveness of \modelname.
We release our codes and models on our project page: \href{https://powerpaint.github.io/}{https://powerpaint.github.io/}. 
\end{abstract}

\section{Introduction}
\label{sec:intro}
Image inpainting aims to fill in user-specified regions in an image with plausible content \cite{Bertalmio2000image}. It has been widely applied in various practical domains, including photo restoration \cite{barnes2009patchmatch,li2017generative,lugmayr2022repaint} and object removal \cite{yu2018generative,suvorov2022lama,pathak2016context}. 
Recently, with the increasing popularity of text-to-image (T2I) models \cite{rombach2022high,saharia2022photorealistic,ramesh2021zero,zhang2023adding}, inpainting has become even more essential. It provides a flexible and interactive approach to mask unsatisfactory regions in generated images and regenerate them for achieving perfect results \cite{wang2023imagen,xie2023smartbrush}.

Despite the significant practical benefits, achieving high-quality versatile image inpainting remains a challenge \cite{yu2018generative,suvorov2022lama,lugmayr2022repaint}. 
Early works focus on context-aware image inpainting, where models are trained by randomly masking a region in an image and reconstructing the original content \cite{pathak2016context,yu2018generative,lugmayr2022repaint}. Such a design aims to incorporate the image context into the inpainted regions, resulting in coherent and visually pleasing completions. However, these models encounter challenges when it comes to synthesizing novel objects since they rely solely on the context to infer the missing content \cite{yu2018generative,suvorov2022lama}.
Recent advancements have seen a shift towards text-guided image inpainting, where a pre-trained T2I model is fine-tuned using masks and text descriptions, resulting in remarkable outcomes in object synthesis \cite{wang2023imagen,xie2023smartbrush,rombach2022high,yang2023uni}. 
However, these approaches introduce a bias that assumes the presence of objects in the masked regions. To remove unwanted objects for a clean background, these models often require extensive prompt engineering or complex workflow. Moreover, these methods remain vulnerable to generating random artifacts that lack coherence with the image context \cite{xie2023smartbrush,wang2023imagen}.

In this paper, we introduce \textbf{\modelname}, the first versatile inpainting model that excels in both text-guided object inpainting and context-aware image inpainting. Our approach capitalizes on the use of distinct learnable task prompts and tailored training strategies for each task, enabling \modelname\ to handle multiple inpainting tasks within a single model. 
Specifically, \modelname\ is built upon a pre-trained T2I diffusion model \cite{rombach2022high}. 
To fine-tune the T2I model for different inpainting tasks, we introduce two learnable task prompts, \pinsert\ and \premove, for text-guided object inpainting and context-aware image inpainting, respectively.
\pinsert\ is optimized by using object bounding boxes as masks and appending \pinsert\ as a suffix to the text description, while \premove\ is optimized with random masks and \premove\ itself as the text prompt.
Through such training, \pinsert\ is able to prompt the \modelname\ to synthesize novel objects based on text descriptions, while using \premove\ can fill in coherent results according to the image context without any additional text hints.
Moreover, the learned task prompt in \modelname\ has effectively captured the intrinsic pattern of the task and can be extended to facilitate powerful object removal.
In particular, existing T2I models employ a classifier-free guidance sampling strategy, where a negative prompt can effectively suppress undesired effects \cite{dhariwal2021diffusion,ho2021classifier}. By leveraging this sampling strategy and designating \premove\ as the positive prompt and \pinsert\ as the negative prompt, \modelname\ effectively prevents the generation of unwanted objects and promotes seamless background filling in the target region, leading to a significant improvement in object removal \cite{rombach2022high}.

To demonstrate the versatility of our task prompts, we further explore a novel prompt interpolation operation for object inpainting, enabling controllable shape-fitting degree to the mask. This task involves balancing text-guided object inpainting in the central region of the mask with context-aware background filling near the periphery. 
During training, we randomly expand object segmentation masks to create inpainting masks and interpolate between \premove\ and a new task prompt, \pshape, based on the expanded area ratio. After training, users can control the shape-fitting degree to the mask by interpolating between \premove\ and \pshape. 
Our main contributions are as follows:
\begin{itemize}
\item To the best of our knowledge, \modelname\ is the first versatile inpainting model that achieves state-of-the-art results in multiple inpainting tasks.
\item We demonstrate the versatility of the task prompts in \modelname, showcasing their capability for object removal by negative prompts and object inpainting with controllable shape-fitting by prompt interpolation.
\item We conducted extensive experiments including both quantitative and qualitative evaluations, to verify the effectiveness of \modelname\ in addressing a wide range of inpainting tasks.
\end{itemize}
\section{Related Work}
\label{sec:related}

\noindent\textbf{Image Inpainting.}
With the significant progress of deep learning, some works have gained remarkable achievements by leveraging generative adversarial networks \cite{goodfellow2014generative,yu2018generative,yu2019free,zeng2022aggregated,Bertalmio2000image,criminisi2004region,barnes2009patchmatch,cao2024leftrefill,zeng2019learning}. These approaches often randomly mask any regions in an image and are optimized to recover the masked region \cite{yu2018generative,pathak2016context,nazeri2019edgeconnect}. Through such optimization, these models are able to fill in the region with content that is coherent with the image context. However, these approaches can not infer new objects from the image context and fail to synthesize novel contents.

Recent advancements have been greatly promoted by text-to-image diffusion models \cite{dhariwal2021diffusion,ho2020denoising,saharia2022photorealistic,ramesh2021zero}. Specifically, SD-Inpainting \cite{rombach2022high} and ControlNet-Inpainting \cite{zhang2023adding} are both built upon the large-scale pre-trained text-to-image model, \ie, Stable Diffusion \cite{rombach2022high}. 
They fine-tune a pre-trained T2I model for inpainting with random masks as the inpainting masks and image captions as the text prompt. 
Despite some good results, these models often suffer from text misalignment and fail to synthesize objects that align with the text prompt. Smartbrush and Imagen Editor propose to address this issue by using paired object-description data for training \cite{xie2023smartbrush,wang2023imagen}. However, these models tend to assume that there are always objects in the missing regions, losing the ability to perform context-aware image inpainting.
We highlight that, through learning different task prompts for different tasks, \modelname\ significantly improves the alignment of text and context, leading to state-of-the-art results in both context-aware image inpainting and text-guided object inpainting. 

\noindent\textbf{Adapting Text-to-Image Models.}
Text-to-image models have achieved remarkable advances recently, showcasing their ability to generate realistic and diverse images based on natural language descriptions \cite{ramesh2021zero,rombach2022high,saharia2022photorealistic}. 
These models have opened up a wide range of applications that utilize their generative power \cite{ruiz2023dreambooth,gal2022image,Kumari_2023_CVPR,sun2024anycontrol,zhang2024pia,tang2024make}. 
One notable example is DreamBooth, which fine-tunes the model to associate specific visual concepts with textual cues, enabling users to create personalized images from text \cite{ruiz2023dreambooth}. Textual Inversion uses a single word vector to encode a unique and novel visual concept, which can then be inverted to generate an image \cite{gal2022image}. Furthermore, Kumari et al. \cite{Kumari_2023_CVPR} propose a method to simultaneously learn multiple visual concepts and seamlessly blend them with existing ones by optimizing a few parameters. 
Instead of learning concept-specific prompts, we propose the utilization of task-specific prompts to guide text-to-image models to achieve various tasks within a single model. Through fine-tuning both the textual embeddings and model parameters, we establish a robust alignment between the task prompts and the desired targets. 

\section{\modelname}
\label{sec:approach} 
To fine-tune a pre-trained text-to-image model for high-quality and versatile inpainting, we introduce three learnable task prompts: \pinsert, \premove, and \pshape, as shown in Figure \ref{fig:FlowChat}.
By incorporating these task prompts along with tailored training strategies, \modelname\ is able to deliver outstanding performance in various inpainting tasks, including text-guided object inpainting, context-aware image inpainting, object removal, and shape-guided object inpainting.

\subsection{Preliminary}
\label{subsec:overall}

\figoverview

\modelname\ is built upon the well-trained text-to-image diffusion model, \ie, Stable Diffusion, which comprises forward and reverse processes \cite{rombach2022high}. 
In the forward process, a noise is added to a clean image $x_0$ in a closed form,
\begin{equation}
    x_t = \sqrt{\bar{\alpha_t}} x_0 + \sqrt{1 - \bar{\alpha_t}} \epsilon, \ \epsilon \sim \mathcal{N}(0, I),
\end{equation} 
where $x_t$ is the noisy image at timestep $t$, and $\bar{\alpha_t}$ denotes the corresponding noise level. 
In the reverse process, a neural network parameterized by $\theta$, denoted as $\epsilon_\theta$, is optimized to predict the added noise $\epsilon_t$. This enables the generation of images by denoising step by step from Gaussian noise. A classical diffusion model is typically optimized by:
\begin{equation}
    \mathcal{L} = \mathbb{E}_{x_0, t, \epsilon_t} \Vert \epsilon_t - \epsilon_\theta(x_t, t) \Vert_2^2.
\end{equation}
To fine-tune Stable Diffusion for inpainting, \modelname\ begins by extending the first convolutional layer of the denoising network $\epsilon_\theta$ with five additional channels specifically designed for the masked image $x_0\odot (1-m)$ and masks $m$. The input to \modelname\ consists of the concatenation of the noisy latent, masked image, and masks, denoted as $x_t'$. 
Additionally, the denoising process can be guided by additional information such as text $y$.
The model is optimized by:
\begin{equation}
\label{eq:loss}
    \mathcal{L} = \mathbb{E}_{x_0, m, t, y, \epsilon_t} \Vert \epsilon_t - \epsilon_\theta(x_t', \tau_\theta(y), t) \Vert_2^2,
\end{equation}
where $\tau_\theta(\cdot)$ is the CLIP text encoder. Importantly, \modelname\ extends the text condition by incorporating learnable task prompts, which serve as guidance for the model to accomplish diverse inpainting tasks. 

\subsection{Learning with Task Prompts}

Context-aware image inpainting and text-guided object inpainting are prominent applications in the field of inpainting, each demanding distinct training strategies for optimal results \cite{pathak2016context, yu2018generative, rombach2022high}.
To seamlessly integrate these two distinct objectives into a unified model, we propose the use of two learnable task prompts dedicated to each task. These task prompts serve as guidance for the model, enabling it to effectively accomplish the desired inpainting targets.

\noindent\textbf{Context-aware Image Inpainting.} 
Context-aware image inpainting aims to fill in the user-specified regions with content that seamlessly integrates with the surrounding image context. Previous studies have shown that training models with random masks and optimizing them to reconstruct the original image yields the best results \cite{pathak2016context, yu2018generative, lugmayr2022repaint}. This training strategy effectively encourages the model to attend to the image context and fill in coherent content.
To achieve this, we introduce a learnable task prompt, denoted as \premove, which serves as the text condition during training. Additionally, we randomly mask the image region as part of the training process. During model fine-tuning, \premove\ is optimized by:
\begin{equation}
\label{eq:lossremove}
    p_{ctxt} =  \underset{p}{\arg\min}  \mathbb{E}_{x_0, m, t, p, \epsilon_t} \Vert \epsilon_t - \epsilon_\theta(x_t', \tau_\theta(p), t) \Vert_2^2,
\end{equation}
where $p$ is randomly initialized as an array of tokens and then used as input to the text encoder. This formulation enables users to seamlessly fill in regions with coherent content without explicitly specifying the desired content.

\noindent\textbf{Text-guided Object Inpainting.}
Synthesizing novel objects that cannot be inferred solely from the image context often requires additional guidance provided by text prompts. Successful approaches in this area have leveraged paired object-caption data during training, allowing the model to generate objects that align with the provided text prompts \cite{rombach2022high, xie2023smartbrush, wang2023imagen}.
To achieve this, we introduce a learnable task prompt, denoted as \pinsert, which serves as the task hint for text-guided object inpainting. Specifically, \pinsert\ shares similar optimization functions as Equation (\ref{eq:lossremove}), but with two differences. First, for a given training image, we utilize the detected object's bounding box as the inpainting mask. Second, we append \pinsert\ as a suffix to the text description of the masked region, which serves as the input to the text encoder.
After training, our model effectively learns to inpaint images based on either the given context or text descriptions. 

\figobjremove
\noindent\textbf{Object Removal.}
\modelname\ can be used for object removal, where users can use a mask to cover the entire object and condition the model on the task prompt \premove\ to fill in coherent content.
However, it becomes more challenging when attempting to remove objects in crowded contexts. As shown in Figure \ref{Object erasure}, even state-of-the-art solutions like Adobe Firefly \cite{adobefirefly2023}, while generating visually pleasing content, tend to synthesize objects within the masked region. We suspect that the inherent network structure, which includes attention layers, leads to the model paying excessive attention to the context. This makes it easier for the model to `copy' information from the crowded context and `paste' it into the masked region, resulting in object synthesis instead of removal.

Fortunately, \premove\ and \pinsert\ can be combined with a powerful classifier-free guidance sampling strategy \cite{ho2021classifier} to achieve effective object removal. This strategy transforms the denoising process into the following form:
\begin{equation}
\begin{aligned}
\widetilde{\epsilon_\theta} =
w \cdot \epsilon_\theta(x_t', \tau_\theta(p_{ctxt}), t) + (1-w) \cdot \epsilon_\theta(x_t', \tau_\theta(p_{obj}), t),
\end{aligned}
\end{equation}
where \premove\ is considered a positive prompt, while \pinsert\ is considered a negative prompt, and $w$ is the guidance scale. The classifier-free guidance strategy works by decreasing the likelihood conditioned on the negative prompt and increasing the likelihood conditioned on the positive prompt for the sample.
With this design, the likelihood of generating objects can be effectively decreased to achieve object removal, as demonstrated in Figure \ref{Object erasure}. This outcome indicates that the task prompts in \modelname\ have successfully captured the patterns associated with different inpainting tasks.


\noindent\textbf{Controllable Shape Guided Object Inpainting.}
In this part, we explore shape-guided object inpainting, where the generated object aligns well with the given mask shape. To achieve this, we introduce a third task prompt, denoted as \pshape, which is trained using precise object segmentation masks and object descriptions, following previous works \cite{xie2023smartbrush}. However, we have noticed that relying solely on \pshape\ can lead the model to overfit the mask shape while disregarding the overall shape of the object. For instance, when provided with the prompt ``a cat" and a square mask, the model may generate cat textures within the square mask without considering the realistic shape of a cat.

\figshapeguided
To address the above limitation and offer users a more reasonable and controllable shape-guided object inpainting, we propose task prompt interpolation.
We start by randomly dilating the object segmentation masks using a convolutional-based dilation operation $D$, which is denoted as,
\begin{equation}
    m' = D(m, k, it)
\end{equation}
where $k$ denotes the kernel size, and $it$ denotes the iteration of dilation.
This generates a set of masks with varying fitting degrees to the object shape. 
For each training mask, we calculate the area ratio, $\alpha$, representing the fitting degree. A larger $\alpha$ indicates a closer fit to the mask shape, while a smaller $\alpha$ indicates a looser fit.
To perform prompt interpolation, we append \pshape\ and \premove\ as suffixes to the text description $y$ and separately input them into the CLIP Text Encoder. This yields two text embeddings. By linearly interpolating these embeddings based on the value of $\alpha$, as shown in Figure ~\ref{fig:shape_insert}, we obtain the final text embedding, which is denoted as:
\begin{equation}
\tau'_{\theta}= (1-\alpha) \cdot \tau_{\theta}(y, p_{ctxt}) + \alpha \cdot \tau_{\theta}(y, p_{shape}).
\end{equation}
After training, users can adjust the value of $\alpha$ to control the fitting degree of the generated objects to the mask shape.

\subsection{Implementation Details}
\label{subsec:imp}
We fine-tune the task prompts in the embedding layer of the CLIP text encoder and the U-Net based on the SD v1.5 model. \modelname\ was trained for 25K iterations on 8 A100 GPUs with a batch size of 1024 and a learning rate of 1e-5. We use the semantic segmentation subset of OpenImage v6\cite{Kuznetsova_2020} as the main dataset for multi-task prompt tuning. In addition, following Smartbrush \cite{xie2023smartbrush}, we use segmentation labels and BLIP captions\cite{li2023blip2} as local text descriptions. Simultaneously, we treat the text-to-image generation task as a special case of inpainting (mask everything), and use the image/text pairs from the LAION-Aesthetics v2 5+\cite{schuhmann2022laion5b} for training. The main task and the text-to-image generation task have probabilities of 80\% and 20\%, respectively, in the training phase.

\section{Experiments}
\label{sec:conclusion}

\textbf{Baselines.}
We select the most recent and competitive inpainting approaches for fair comparisons. We list them with brief introductions below: 
\begin{itemize}
\item \textbf{LaMa} \cite{suvorov2022lama} is built upon a generative adversarial network \cite{goodfellow2014generative} and achieves state-of-the-art in large mask inpainting. 
\item \textbf{LDM-Inpainting} \cite{rombach2022high} is finetuned from a text-to-image latent diffusion model for inpainting without text prompt.
\item \textbf{Blended Diffusion} \cite{avrahami2022blended} achieves text-guided inpainting by leveraging a language-image model (CLIP) \cite{radford2021clip}.
\item \textbf{Stable Diffusion} \cite{rombach2022high} achieves text-guided inpainting by blending the unmasked latent in each denoising step. 
\item \textbf{SD-Inpainting} \cite{rombach2022high} fine-tuned Stable Diffusion with random masks and image caption for inpainting.
\item \textbf{CN-Inpainting} \cite{zhang2023adding} controls Stable Diffusion for inpainting by a controlnet that encodes masked images. 
\item \textbf{SmartBrush} \cite{xie2023smartbrush} fine-tuned Stable Diffusion with object masks of varying granularity and localized text descriptions. 
\end{itemize}

\noindent\textbf{Evaluation Benchmarks.}
To make fair comparisons with SOTA approaches, we adopt the most commonly-used datasets for inpainting following previous works \cite{suvorov2022lama,xie2023smartbrush,rombach2022high}. 
First, we evaluate object inpainting on \textbf{OpenImages} \cite{Kuznetsova_2020} and \textbf{MSCOCO} \cite{lin2014microsoft} datasets, following Smartbrush \cite{xie2023smartbrush}. Each dataset comprises approximately 10K images and corresponding masks.
Second, we evaluate context-aware image inpainting on \textbf{Places2} \cite{zhou2017places}. We sample 10k images from the test set of Places2 and generate random masks as inpainting masks following Rombach et al. \cite{suvorov2022lama,rombach2022high}. In this setting, there is no text prompt provided for evaluation and the inpainting model should fill in regions according to image contexts. 
Finally, we evaluate the performance of image outpainting without text prompt on 10K images from \textbf{Flickr-Scenery}, which are the most representative and natural use-cases of outpainting, following Cheng \etal \cite{yang2019very,teterwak2019boundless,cheng2022inout}. 

\noindent\textbf{Evaluation Metrics.}
We use five Fréchet Inception Distance (FID) \cite{heusel2017fid}, Local-FID \cite{xie2023smartbrush}, CLIP Score \cite{radford2021clip}, LPIPS \cite{zhang2018unreasonable}, and aesthetic score \cite{schuhmann2022laion5b} as numeric metrics for different inpainting tasks, following common settings \cite{xie2023smartbrush,wang2023imagen,suvorov2022lama,rombach2022high,cheng2022inout}. 
Specifically, we use FID and Local-FID for global and local image visual quality. 
The CLIP Score is used in text-guided object inpainting to evaluate the alignment of generated visual content with the text prompt. 
Since context-aware image inpainting aims at recovering the randomly masked regions according to image contexts, we use the original image as ground truths and use LPIPS to evaluate the reconstruction performance. 
Finally, we introduce an aesthetic score for outpainting, which aims to evaluate the extended content for pleasing scenery.

\tabbbox
\tabobj

\tabinpaint
\taboutpaint

\subsection{Comparisons with State-of-the-Art}
\label{subsec:expsota}

\noindent\textbf{Quantitative Comparisons.}
We report quantitative evaluation on various inpainting benchmarks following previous works \cite{suvorov2022lama,xie2023smartbrush}. 
For text-guided object inpainting and shape-guided object inpainting, we use bounding box masks and object layout masks for testing on OpenImages \cite{Kuznetsova_2020} and MSCOCO \cite{lin2014microsoft}, respectively. 
The results in Table \ref{tab:bbox} and Table \ref{tab:objmask} demonstrate that \modelname\ is able to generate realistic and diverse images that satisfy both the text and shape constraints. In particular, \modelname\ achieves state-of-the-art in terms of both visual quality and text alignment for object inpainting. 

For context-aware image inpainting, we include text-free inpainting models, \ie, LaMa \cite{suvorov2022lama} and LDM-Inpaint \cite{rombach2022high}, and the strongest baselines in Table \ref{tab:bbox} and \ref{tab:objmask}, \ie,  SD-Inpainting for further comparison. 
The quantitative results shown in Table \ref{tab:bgfill} demonstrate that \modelname\, guided by a task-specific prompt, outperforms the baseline in effectively filling missing regions while better matching the image context. 
To conduct a thorough comparison, we use a default text prompt of "background" and "scenery" with SD-Inpainting, and compare it with \modelname\ using \pinsert\ as a negative prompt. Notably, we observed that using \pinsert\ as a negative prompt effectively reduces the generation of random artifacts and preserves a coherent background that aligns with the image context. This improvement leads to significantly improved inpainting results.

We report the quantitative comparison for image outpainting in \cref{tab:outpaint}. Since image outpainting is often required to extend the image with content that is both aesthetically pleasing and coherent, we employed FID and aesthetic scores for its quantitative evaluation. As indicated in Table \ref{tab:outpaint}, our model demonstrates superior image and aesthetic quality compared to the baseline models.

\figinsert
\figinpaint
\figoutpaint

\noindent\textbf{Qualitative Comparisons.}
The qualitative comparison in Figure \ref{fig:biginsert}, \ref{fig:bigcontext} and \ref{fig:bigoutpaint} show that our model has achieved state-of-the-art performance in text-guided object inpainting, context-aware image inpainting, and outpainting. 
For text-guided object inpainting, existing models often fail to synthesize objects that are faithful to the text prompt. For example, in the fourth case, CN-Inpainting and SD-Inpainting are hard to generate trousers in the region and can only fill the region with backgrounds. \modelname\ is able to synthesize high-fidelity objects according to the text prompt with both bounding box masks and object layout masks.  
For context-aware image inpainting and outpainting, our model outperforms both text-free and text-based inpainting models significantly. Taking the second case in Figure \ref{fig:bigcontext} as an example, LaMa tends to synthesize blurry results due to its limited generation capacity, while Adobe Firefly \cite{adobefirefly2023} tends to generate random objects in the region, which goes against the users' intention. 

\noindent\textbf{User study.}
\tabuser
We conducted user studies for a more comprehensive comparison. Specifically, we deliver three groups of user studies for text-guided object inpainting, object removal, and outpainting, respectively. 
For each group, we randomly sample test images and show the inpainting results to volunteers anonymously.
To ensure stable and convincing results with minimal user effort, we specifically selected the two strongest baselines for each user study group according to their quantitative and qualitative results, instead of considering all baselines.

In each trial, we introduce different inpainting tasks to the volunteers and ask them to choose the most satisfying results per different targets. Specifically, for the object inpainting task, we conducted a more detailed investigation into user preferences, examining aspects including shape, text alignment, and realism.
We have collected 2,995 valid votes and conclude the results in Table \ref{tab:humaneval}. The results show that our model is preferred in all three tasks.

\subsection{Ablation Study}

\noindent\textbf{Effectiveness of Learnable Task Prompt.}
To verify the effectiveness of task prompt learning, we compare our model with the variant of tuning with unlearnable rare identifiers \cite{ruiz2023dreambooth}. 
In this variant, we use different rare identifiers to denote different tasks with the same training strategies as \modelname.  
The quantitative comparison in Table \ref{tab:ablation_guide} shows that a learnable task prompt can be better compatible with the description and conditions (\ie, masked images and masks) for different inpainting targets, leading to better results.

\noindent\textbf{Single unified VS task-specific.}
We trained separate task-specific models for text-guided object inpainting, shape-guided object inpainting, and context-aware image inpainting. 
The quantitative results in \cref{tab:task-specific} show that PowerPaint, as a versatile model, achieves comparable performance to the task-specific models, sometimes even better results. 
This indicates the effectiveness of incorporating task prompts in a unified model without compromising performance. 

\noindent\textbf{Fine-training dataset.}
To alleviate concerns regarding inconsistencies in pre-training datasets, we conducted additional experiments by fine-tuning the SD-Inpainting [25] model on the fine-tuning dataset utilized by PowerPaint, namely, OpenImages [15] and LAION Aesthetics v2 5+ [28]. Our results demonstrate marginal improvements over the baseline when fine-tuning on the same dataset, consistent with the observations made by Smartbrush \cite{xie2023smartbrush}.
\tababguide
\tabSDinptask

\subsection{Applications and Limitations}
\figremove

\noindent\textbf{Object Removal.}
We find it challenging to remove objects from crowded image contexts for inpainting models based on diffusion model, which often copies context objects into regions due to the intrinsic network structure (\ie, self-attention layers). We show in Figure \ref{fig:obj_remove} that, combined with classifier-free guidance strategy, our model uses \pinsert\ as a negative prompt so that it can prevent generating objects in regions for effective object removal.

\noindent\textbf{Shape-Guided Object Inpainting.}
Given a mask, \modelname\ enables users to control the fitting degree to the mask shape by adjusting the interpolation of two leaned task prompts, i.e., \premove\ and \pshape. 
Results in Figure \ref{fig:shape_guided} show that \modelname\ can synthesize high-fidelity results that are faithful to both the mask shape and text prompt.

\noindent\textbf{Limitations.}
First, the visual quality can be constrained by the capabilities of the underlying text-to-image diffusion model. 
Second, in the case of shape-guided object inpainting, achieving a fitting degree with extremely small values is challenging. This limitation stems from the fact that there are few instances in which the object occupies a very small area during training.

\figshape  
\section{Conclusions}
\label{sec:conclu}
We present \modelname\ as a versatile inpainting model that achieves state-of-the-art performance across multiple inpainting tasks. We attribute the success of \modelname\ to the utilization of task prompts and tailored optimal training strategies. We conduct extensive experiments and applications to verify the effectiveness of \modelname\ and the versatility of the task prompt through applications of removal and object inpainting with controllable shape-fitting. 

\section{Appendix} 
\appendix
We have included our codes, models, and supplementary material as part of our submission. This material provides additional results of the qualitative comparison with state-of-the-art approaches in Section \ref{sec:supp_exp}.
Furthermore, we present more application results in Section \ref{sec:supp_app}. Specifically, in Section \ref{subsec:supp_removal}, we demonstrate object removal. In Section \ref{subsec:supp_shape}, we showcase shape-guided object inpainting with a controllable fitting degree. Additionally, in Section \ref{subsec:supp_control}, we discuss the combination of our approach with ControlNet \cite{zhang2023adding}.

\section{Qualitative Comparisons}
\label{sec:supp_exp}
We present a comprehensive comparison of \modelname\ with state-of-the-art methods in various inpainting tasks. These tasks include text-guided object inpainting, shape-guided object inpainting, context-aware image inpainting, and image outpainting. To ensure fairness, the results we showcase are randomly sampled, avoiding any cherry-picking to provide a more accurate demonstration.

\begin{figure*}[htbp]
    \centering
    \includegraphics[width=0.97\textwidth]{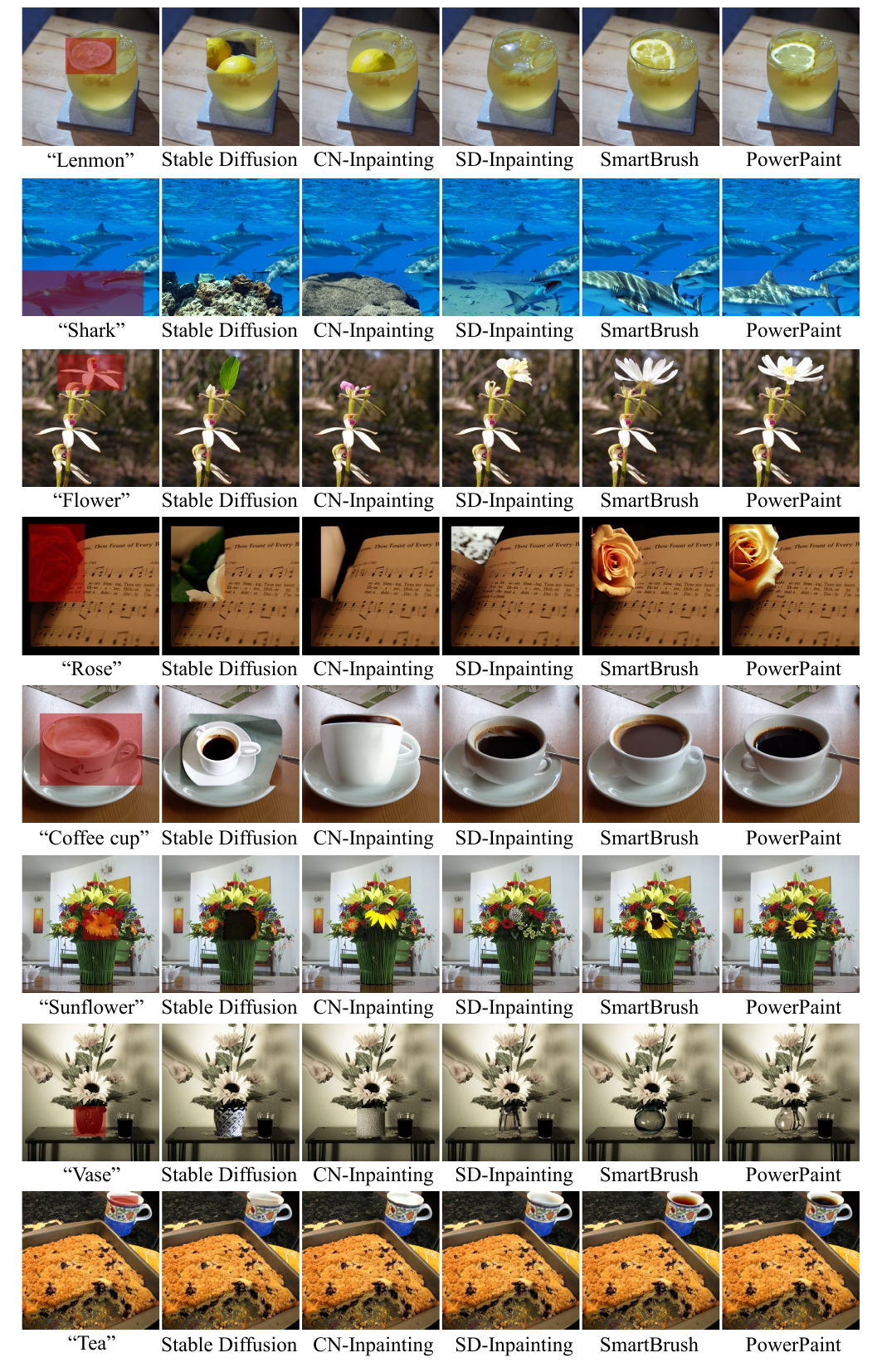}
    \caption{\textbf{Text-guided object inpainting.} We compare \modelname\ with Stable Diffusion \cite{rombach2022high}, CN-Inpainting \cite{zhang2023adding}, SD-Inpainting \cite{rombach2022high} and SmartBrush\cite{xie2023smartbrush}. \modelname\ shows state-of-the-art text alignment and visual quality. [Best viewed with zoom-in in color]}
    \label{fig:sup_object1}
\end{figure*}

\begin{figure*}[htbp]
    \centering
    \includegraphics[width=0.97\textwidth]{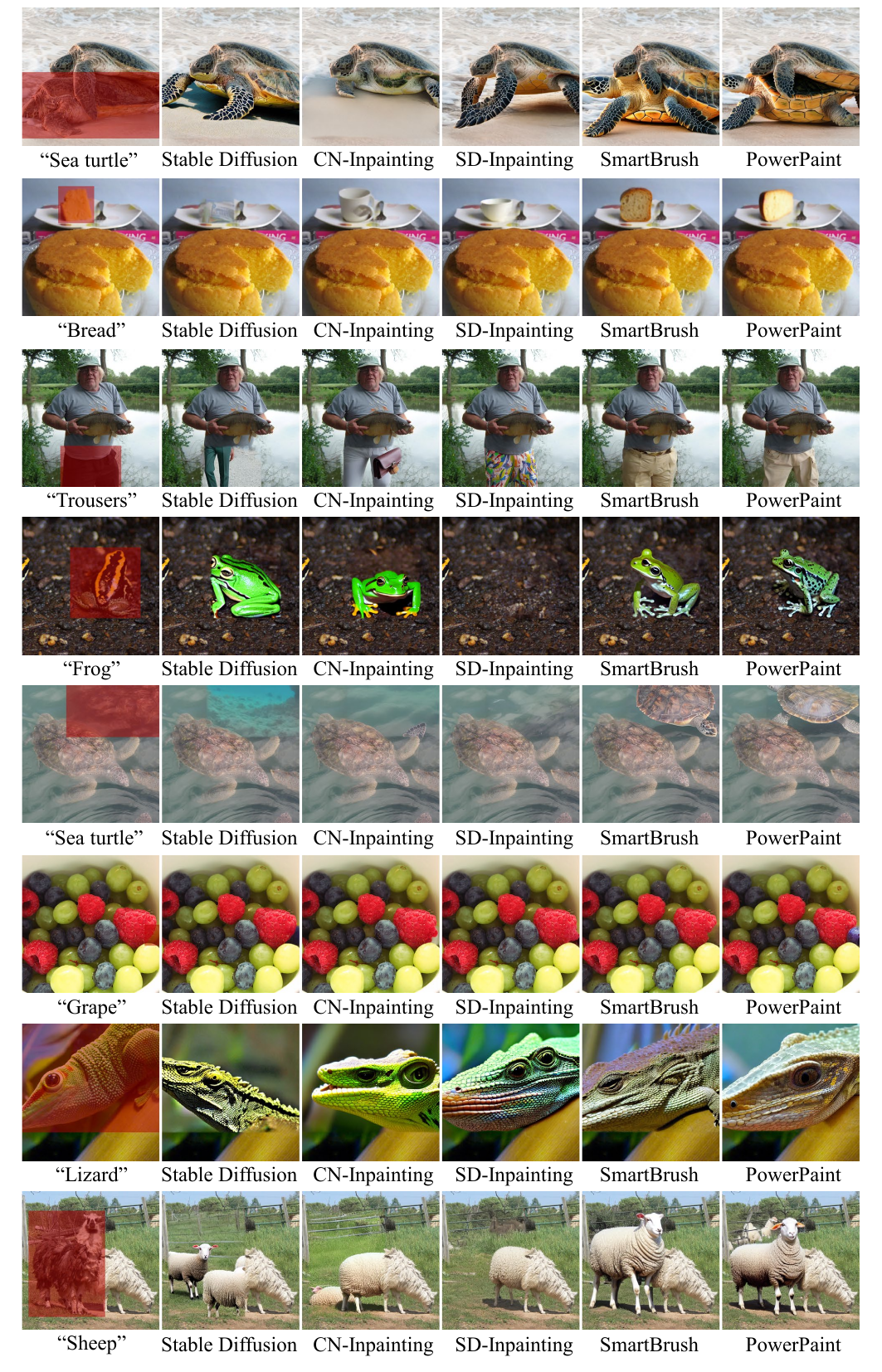}
    \caption{\textbf{Text-guided object inpainting.} We compare \modelname\ with Stable Diffusion \cite{rombach2022high}, CN-Inpainting \cite{zhang2023adding}, SD-Inpainting \cite{rombach2022high} and SmartBrush\cite{xie2023smartbrush}. \modelname\ shows state-of-the-art text alignment and visual quality. [Best viewed with zoom-in in color]}
    \label{fig:sup_object2}
\end{figure*}

\paragraph{\normalfont \textbf{Text-guided object inpainting.}}
In addition to Fig. 5 in the main paper, we provide additional qualitative results of text-guided object inpainting in \cref{fig:sup_object1} and \cref{fig:sup_object2} for a more comprehensive comparison. For these comparisons, we carefully selected the most recent and competitive baselines, including Stable Diffusion \cite{rombach2022high}, CN-Inpainting \cite{zhang2023adding}, SD-Inpainting \cite{rombach2022high}, and SmartBrush \cite{xie2023smartbrush}.

As shown in \cref{fig:sup_object1} and \cref{fig:sup_object2}, the first column displays the input image, the prompt used for inpainting, and the target inpainting region marked in red. Subsequently, we compare the inpainting results generated by state-of-the-art approaches and \modelname.
Stable Diffusion, as demonstrated in the results, utilizes a method introduced in SDEdit \cite{meng2021sdedit} to extend a text-to-image model for image inpainting. While Stable Diffusion can occasionally fill in regions based on the text prompt, it struggles to generate content coherent with the image context.
CN-Inpainting and SD-Inpainting, fine-tuned for inpainting based on Stable Diffusion, exhibit more coherent inpainting results. However, during the fine-tuning process, these methods use random masks and image captions for training, which often results in misalignment with the prompt describing the inpainting region.
SmartBrush, a model specifically trained for text-guided object inpainting, is included for comparison. We can observe that \modelname\ achieves comparable object inpainting results that effectively match the text descriptions and input images. Notably, \modelname\ is a versatile inpainting model that also excels at object removal, a task that SmartBrush struggles to accomplish.

\paragraph{\normalfont \textbf{Shape-guided object inpainting.}}
In addition to Fig. 5 in the main paper, we provide additional object inpainting results in \cref{fig:biginsert} by using exact object layouts as inpainting masks. Similar to the results of text-guided object inpainting using bounding boxes as inpainting masks in \cref{fig:sup_object1} and \cref{fig:sup_object2}, Stable Diffusion struggles to fill in inpainting regions with content that matches the image context. CN-Inpainting and SD-Inpainting show better results in completing user-specified regions with coherent content. However, these methods may fail to synthesize content that satisfies the prompt description. For example, in the third case in \cref{fig:biginsert}, both CN-Inpainting and SD-Inpainting fail to generate a sea turtle in the results.

Both SmartBrush and \modelname, on the other hand, excel at generating pleasing results that not only match the text prompt and image context but also align well with free-form object layouts. This success can be attributed to their optimal training strategy for object inpainting, which incorporates object masks and object descriptions during training. It is important to highlight the superiority of \modelname\ as a versatile inpainting model, achieving comparable performance in text-guided and shape-guided object inpainting tasks, even when compared to specially-trained object inpainting models like SmartBrush.

\begin{figure*}[htbp]
    \centering
    \includegraphics[width=0.97\textwidth]{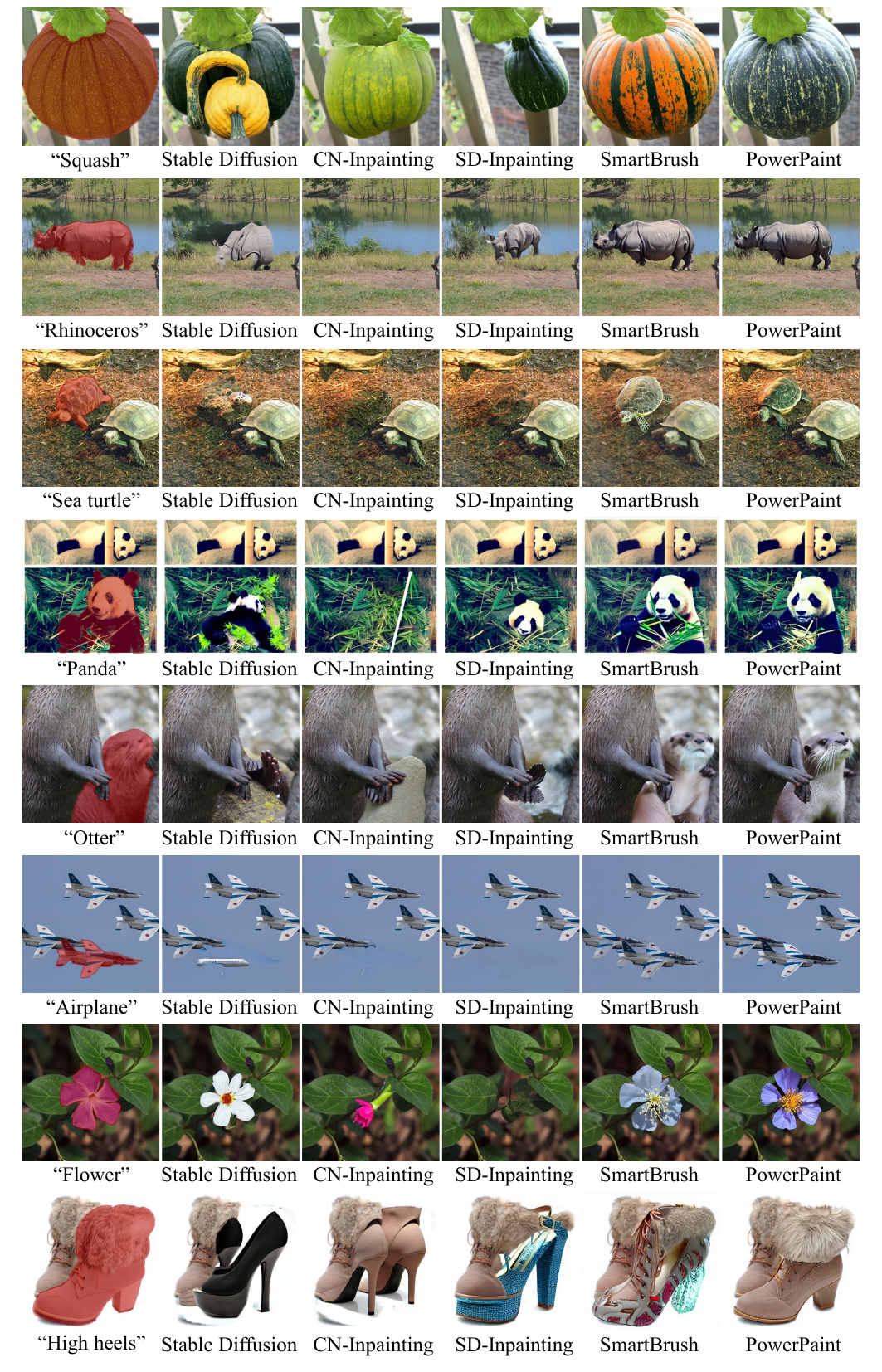}
    \caption{\textbf{Shape-guided object inpainting.} We use exact object layout masks as inpainting masks. We compare \modelname\ with Stable Diffusion \cite{rombach2022high}, CN-Inpainting \cite{zhang2023adding}, SD-Inpainting \cite{rombach2022high} and SmartBrush \cite{xie2023smartbrush}.\modelname\ shows state-of-the-art text alignment and shape alignment. [Best viewed with zoom-in in color]}
    \label{fig:biginsert}
\end{figure*}

\begin{figure*}[htbp]
    \centering
    \includegraphics[width=0.97\textwidth]{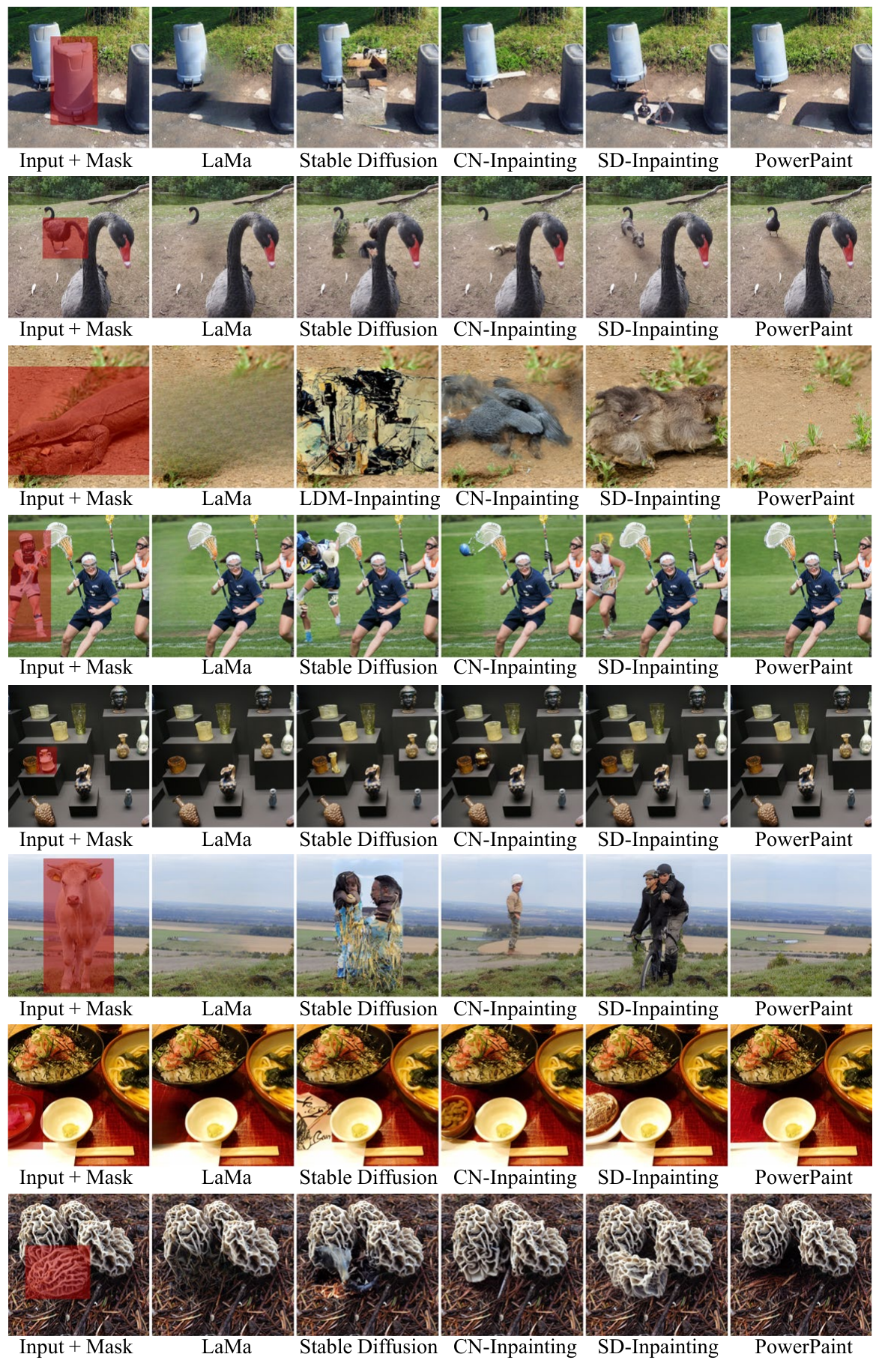}
    \caption{\textbf{Context-aware image inpainting.} Users do not need input text prompts for context-aware image inpainting. We compare \modelname\ with LaMa \cite{suvorov2022lama}, Stable Diffusion \cite{rombach2022high}, CN-Inpainting \cite{zhang2023adding} and SD-Inpainting \cite{rombach2022high}. \modelname\ can synthesize high-quality and context-aware results. [Best viewed with zoom-in in color]}
    \label{fig:sup_remove}
\end{figure*}

\paragraph{\normalfont \textbf{Context-aware image inpainting.}}
In addition to Fig. 6 in the main paper, we provide an additional qualitative comparison with state-of-the-art approaches in \cref{fig:sup_remove}.
In the case of context-aware image inpainting, users do not need to provide any prompt, and the inpainting model is expected to fill in the region with reasonable results that are coherent with the image context. This technique is often used in automatic image restoration or batch object removal.
We carefully selected the baselines for context-aware image inpainting for comparison, including LaMa \cite{suvorov2022lama}, Stable Diffusion \cite{rombach2022high}, CN-Inpainting \cite{zhang2023adding}, and SD-Inpainting.

In \cref{fig:sup_remove}, we observe that LaMa, an inpainting model based on a generative adversarial network, has limitations in synthesizing realistic and visually pleasing results. On the other hand, Stable Diffusion, CN-Inpainting, and SD-Inpainting produce more visually appealing results, leveraging the generative capabilities of large diffusion models. However, these methods often rely on prompt engineering to achieve satisfactory outcomes, and in the absence of detailed prompts, they may introduce random artifacts into the results.
In contrast, \modelname\ stands out among existing methods by generating realistic and coherent content that aligns with the image context without any text hints. 
For example, in the second case of \cref{fig:sup_remove}, \modelname\ successfully completes the black goose by considering the shape and context of the goose's neck around the inpainting mask. In the third case of \cref{fig:sup_remove}, where a significant portion of the image is occluded, \modelname\ is able to produce a coherent completion with natural textures.

\begin{figure*}[htbp]
    \centering
    \includegraphics[width=0.97\textwidth]{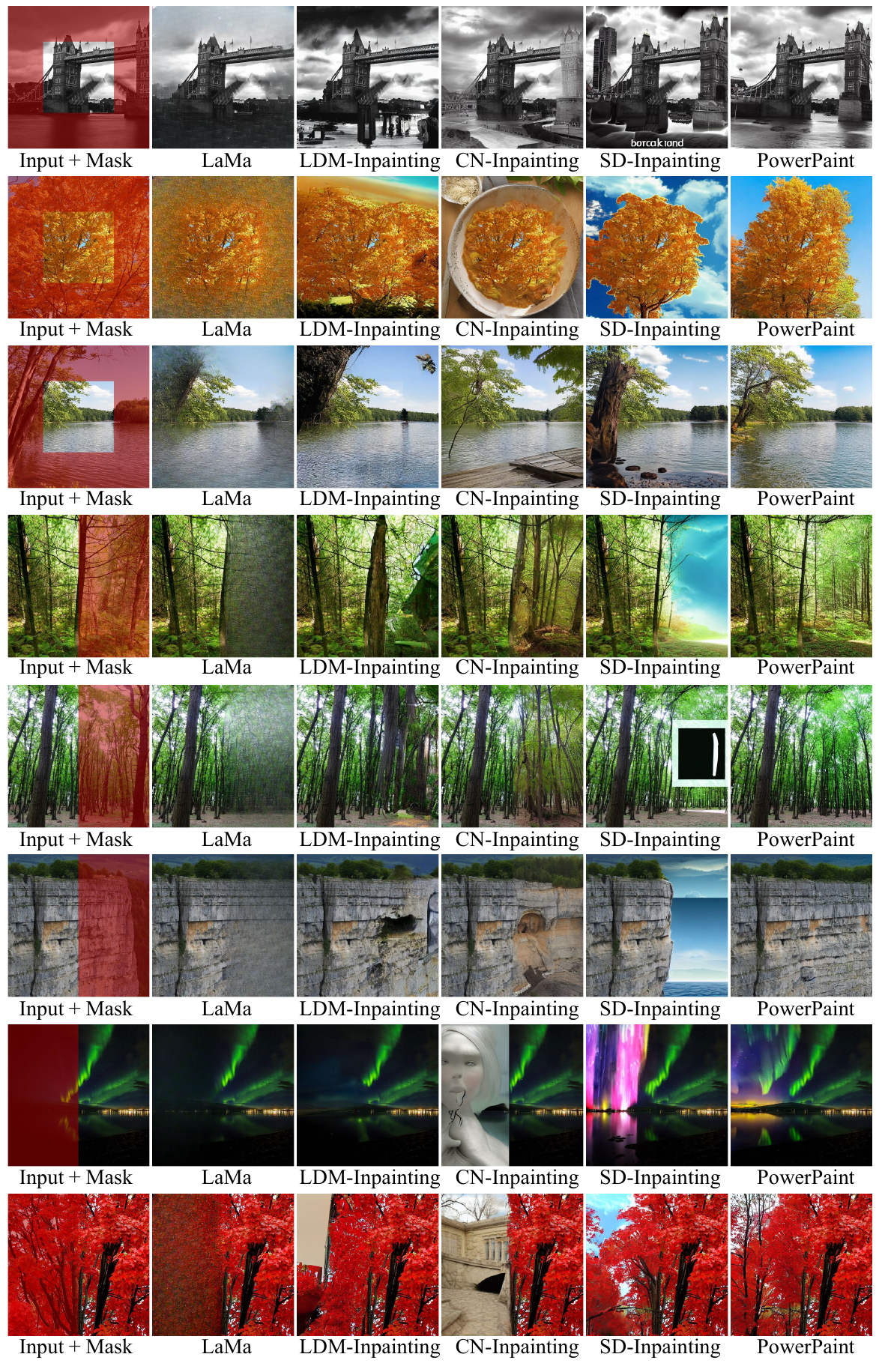}
    \caption{\textbf{Image outpainting.} We compare \modelname\ with LaMa \cite{suvorov2022lama}, LDM-Inpainting \cite{rombach2022high}, CN-Inpainting \cite{zhang2023adding} and SD-Inpainting \cite{rombach2022high} with various outpainting masks. \modelname\ is able to expand the input image with much more reasonable and visually pleasing results. [Best viewed with zoom-in in color]}
    \label{fig:sup_outpaint}
\end{figure*}

\paragraph{\normalfont \textbf{Image outpainting.}}
With the increasing demand for adapting image or video content to different platforms, such as portrait mode in TikTok or landscape mode on a laptop, image outpainting has become increasingly important. The goal of image outpainting is to expand the boundaries of an image with realistic and coherent content that matches the image's context. In addition to the results presented in Figure 7 of the main paper, we provide additional qualitative comparisons with state-of-the-art approaches in Figure \ref{fig:sup_outpaint}. We compare \modelname\ with LaMa \cite{suvorov2022lama}, LDM-Inpainting \cite{rombach2022high}, CN-Inpainting \cite{zhang2023adding}, and SD-Inpainting \cite{rombach2022high} in terms of their performance in image outpainting, which represents the state-of-the-art inpainting techniques.

As shown in Figure \ref{fig:sup_outpaint}, we evaluate image outpainting using three types of outpainting masks. LaMa struggles to extend the image context with large inpainting masks and often produces unclear textures due to its limited generative capacity. LDM-Inpainting, CN-Inpainting, and SD-Inpainting, on the other hand, are capable of generating pleasing results in most cases, due to their powerful generative capacity by the large pre-trained diffusion models. However, these methods sometimes overlook the image context and generate random artifacts, as demonstrated in the seventh case in Figure \ref{fig:sup_outpaint}. In contrast, \modelname\, as a high-quality and versatile image inpainting model, is capable of extending the image context with visually pleasing and globally coherent content without the need for prompt engineering.

\section{Application Results}
\label{sec:supp_app}

In addition to the results presented in Section 4.4 of the main paper, this section provides additional results on various inpainting applications using \modelname. Specifically, we explore object removal in \cref{subsec:supp_removal}, shape-guided object inpainting with an adjustable fitting degree in \cref{subsec:supp_control}, and the integration of \modelname\ with ControlNet \cite{zhang2023adding} in \cref{subsec:supp_control}.

\begin{figure*}[htbp]
    \centering
    \includegraphics[width=\textwidth]{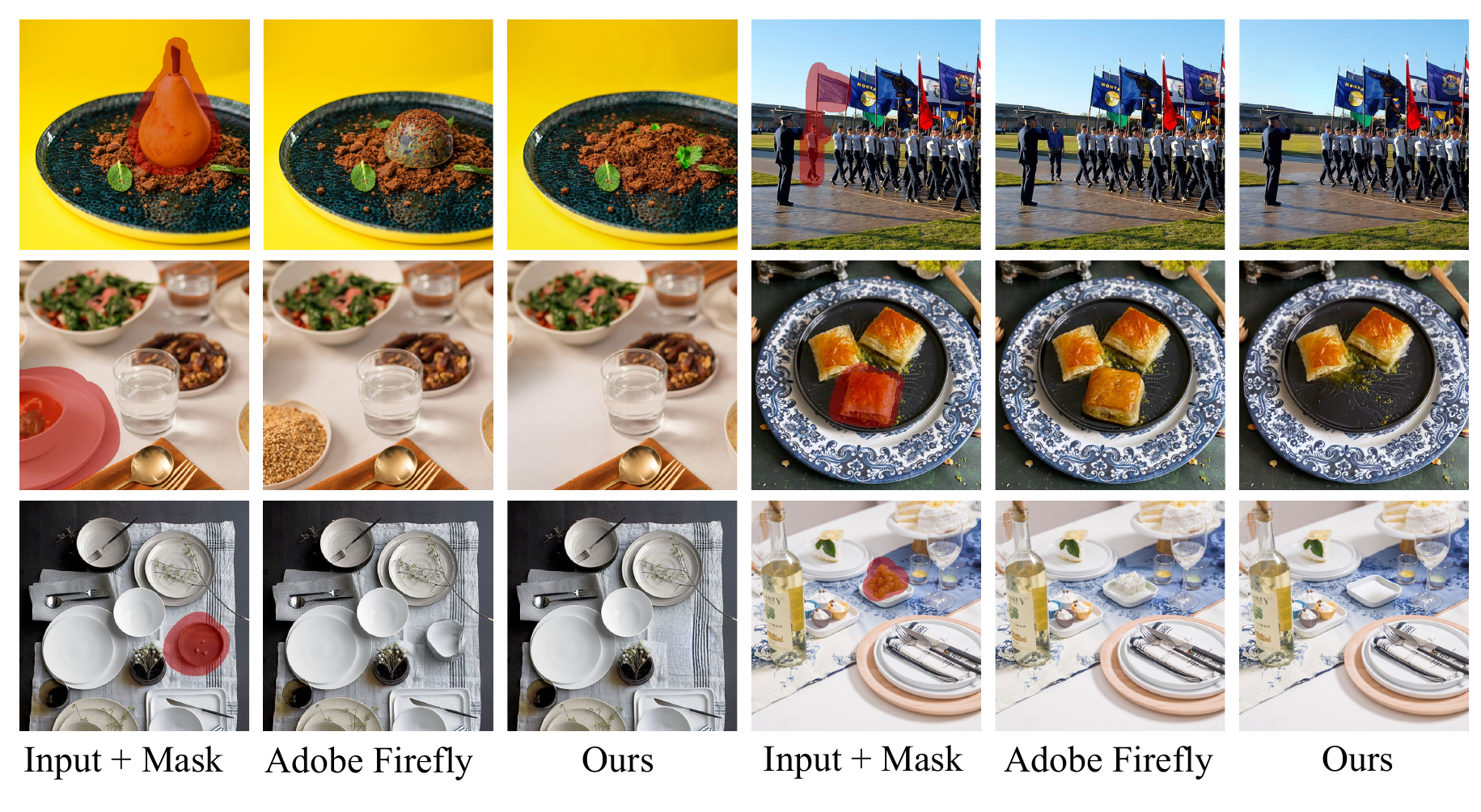}
    \caption{\textbf{Object removal}. We compare \modelname\ with Adobe Firefly \cite{adobefirefly2023}, a commercial product likely based on a large text-to-image model. Following Adobe Firefly's guidelines, we utilize their tool for object removal and find that \modelname\ outperforms it, demonstrating superior results. [Best viewed with zoom-in in color]}
    \label{fig:remove_supp}
\end{figure*}

\subsection{Object Removal}
\label{subsec:supp_removal}
In addition to Figure 8 in the main paper, we provide further comparison results with Adobe Firefly\cite{adobefirefly2023} for object removal in Figure \ref{fig:remove_supp} to address concerns about cherry-picking. As depicted in Figure \ref{fig:remove_supp}, removing objects from crowded image contexts poses a challenge. We suspect that the network's inherent structure, particularly the attention layers, tends to focus excessively on the context and inadvertently copy content from the surrounding areas. Consequently, Adobe Firefly often synthesizes random objects in the inpainting regions.

With the use of learnable task prompts, \modelname\ has successfully captured the distinctive patterns associated with various inpainting tasks. Specifically, the \pinsert\ prompt has learned to generate objects in the masked regions, while the \premove\ prompt has learned to focus on synthesizing content that aligns with the image context. Through the classifier-free guidance sampling strategy \cite{ho2021classifier}, \modelname\ designates \premove\ as the positive prompt and \pinsert\ as the negative prompt. This encourages the model to generate coherent content while avoiding the generation of new objects, resulting in effective object removal.

\subsection{Controllable Shape-Guided Object Inpainting}
\label{subsec:supp_shape}
\begin{figure*}[htbp]
    \centering
    \includegraphics[width=\textwidth]{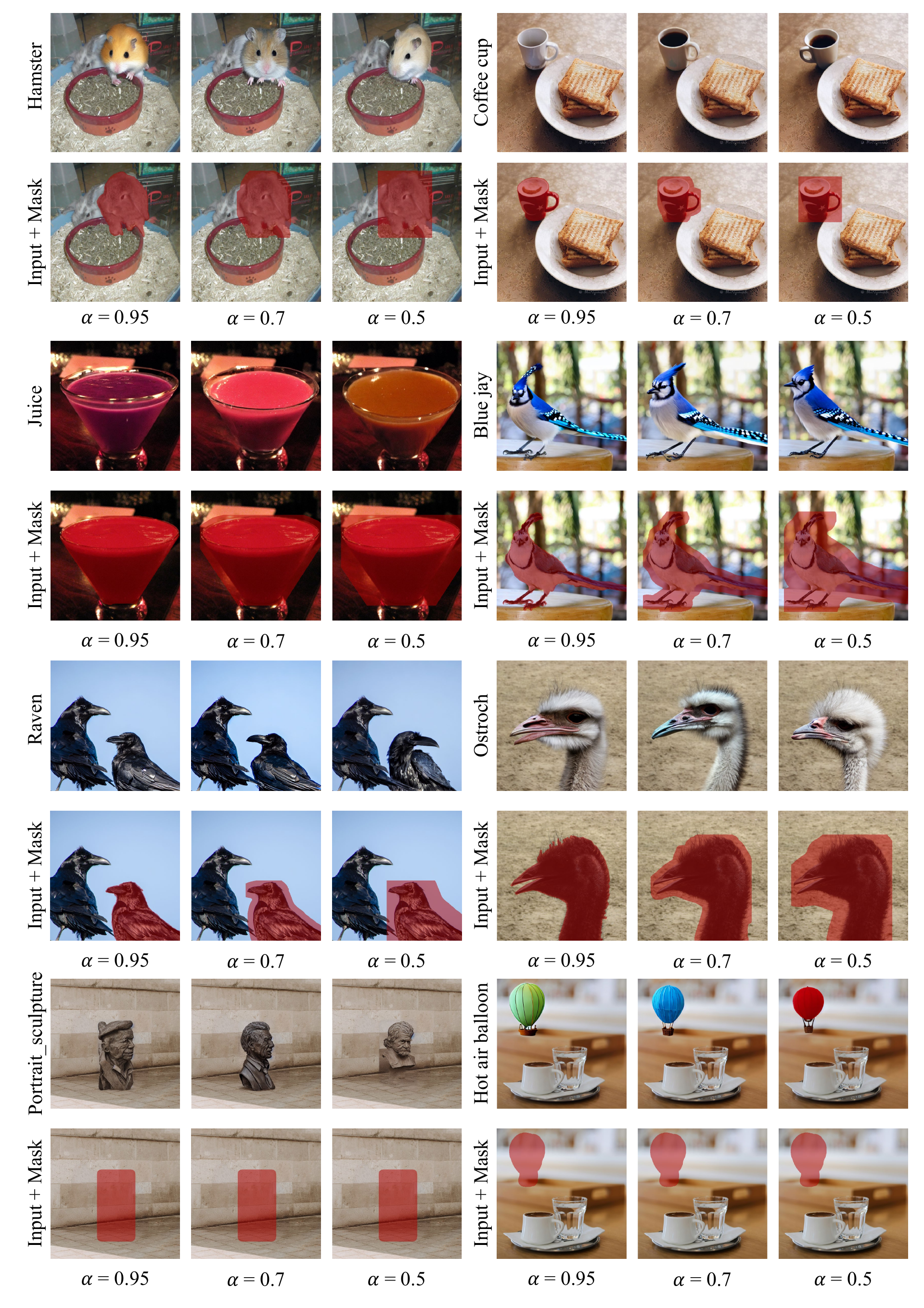}
    \caption{\textbf{Controllable shape-guided object inpainting.} Users have the flexibility to synthesize objects with precise shapes by providing accurate object layouts with a high fitting degree, such as $\alpha=0.95$. Alternatively, they can utilize a coarse object inpainting mask (e.g., a bounding box) and set a relatively lower value for the shape-fitting degree, such as $\alpha=0.5$, to fill in an object with a reasonably plausible shape. 
    [Best viewed with zoom-in in color]}
    \label{fig:shape_supp}
\end{figure*}

In addition to Figure 9 in the main paper, we provide additional visual results for controllable shape-guided object inpainting in Figure \ref{fig:shape_supp}. \modelname\ demonstrates the ability to interpolate between the \premove\ and \pshape\ prompts, allowing for a trade-off between context-aware image inpainting around the contours of the inpainting mask and text-guided object inpainting in the center of the mask. 

As depicted in Figure \ref{fig:shape_supp}, when provided with an accurate object layout and a high value for the shape fitting degree, such as $\alpha=0.95$, \modelname\ synthesizes the object precisely according to the text prompt and the shape of the inpainting mask. Conversely, when given a rough inpainting mask (e.g., a bounding box) and a lower value for the shape fitting degree, such as $\alpha=0.5$, \modelname\ generates an object with a reasonable shape without excessively conforming to the shape of the inpainting mask. The results demonstrate that PowerPaint faithfully adheres to the shape of the inpainting mask, the text prompt, and the desired fitting degree, resulting in realistic and controllable inpainting outputs.

\subsection{\modelname\ with ControlNet}
\label{subsec:supp_control}
\begin{figure*}[htbp]
    \centering
    \includegraphics[width=0.9\textwidth]{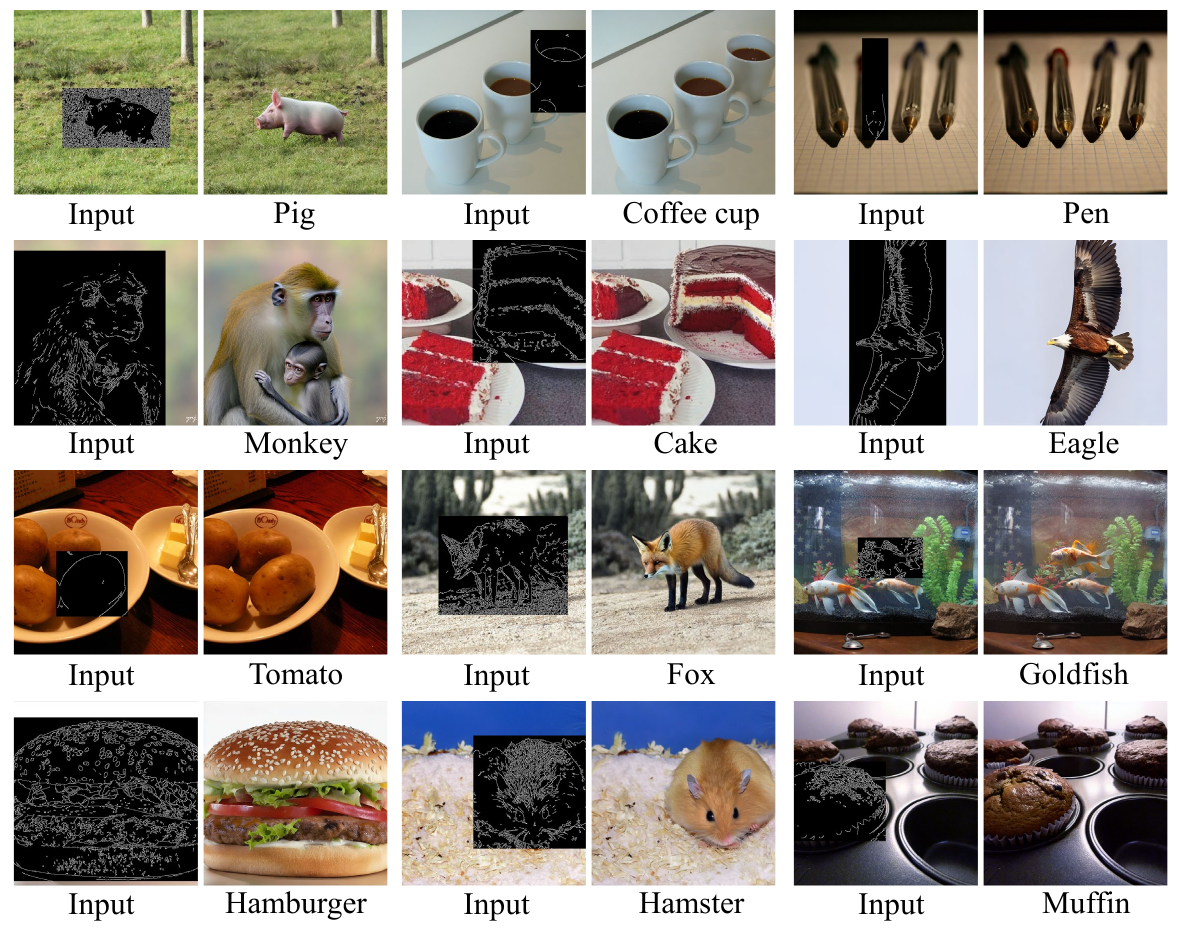}
    \caption{
    Visual results of \modelname\ with the ControlNet conditioned on canny.}
    \label{fig:sup_canny}
\end{figure*}

\begin{figure*}[htbp]
    \centering
    \includegraphics[width=0.9\textwidth]{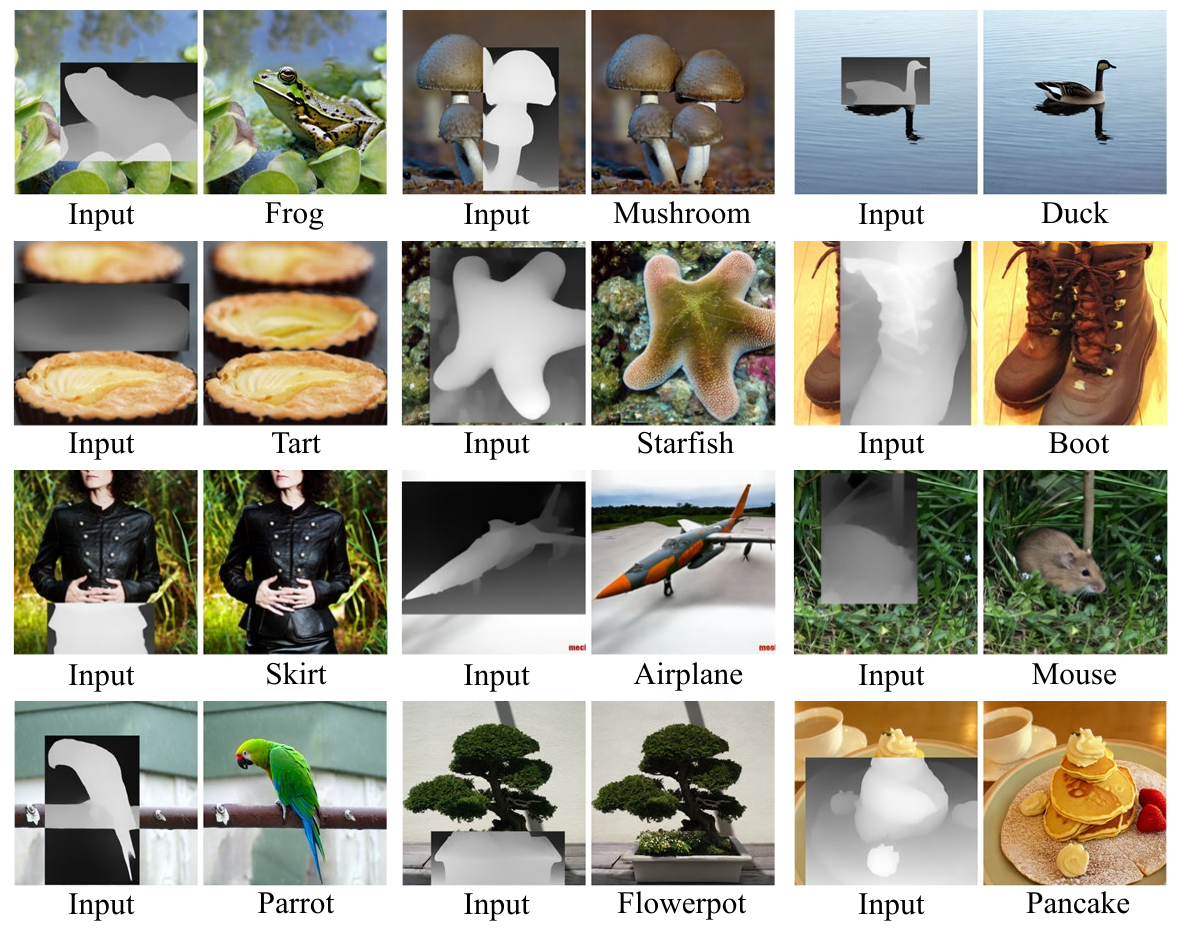}
    \caption{Visual results of \modelname\ with the ControlNet conditioned on depth.}
    \label{fig:sup_depth}
\end{figure*}

\begin{figure*}[htbp]
    \centering
    \includegraphics[width=0.9\textwidth]{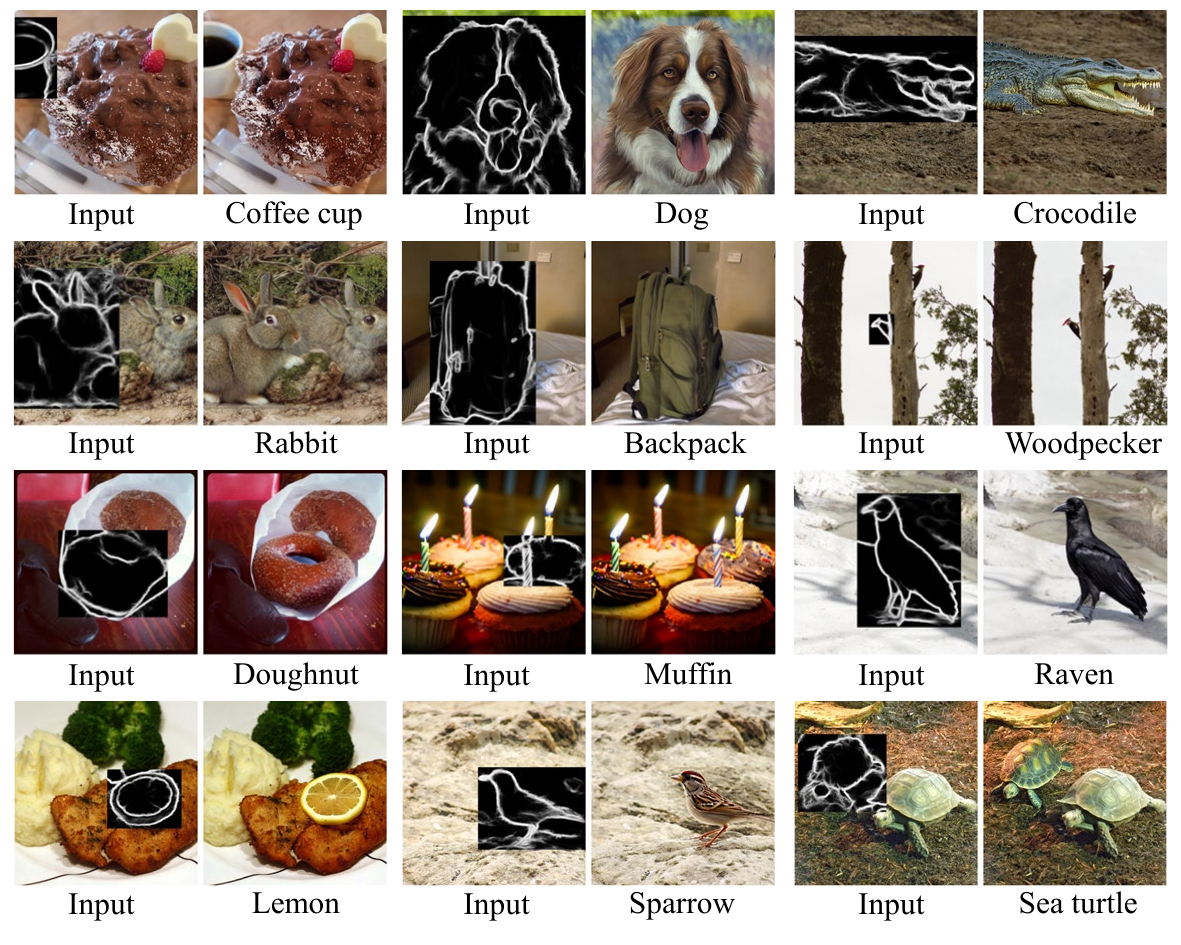}
    \caption{Visual results of \modelname\ with the ControlNet conditioned on hed.}
    \label{fig:sup_hed}
\end{figure*}

\begin{figure*}[htbp]
    \centering
    \includegraphics[width=0.9\textwidth]{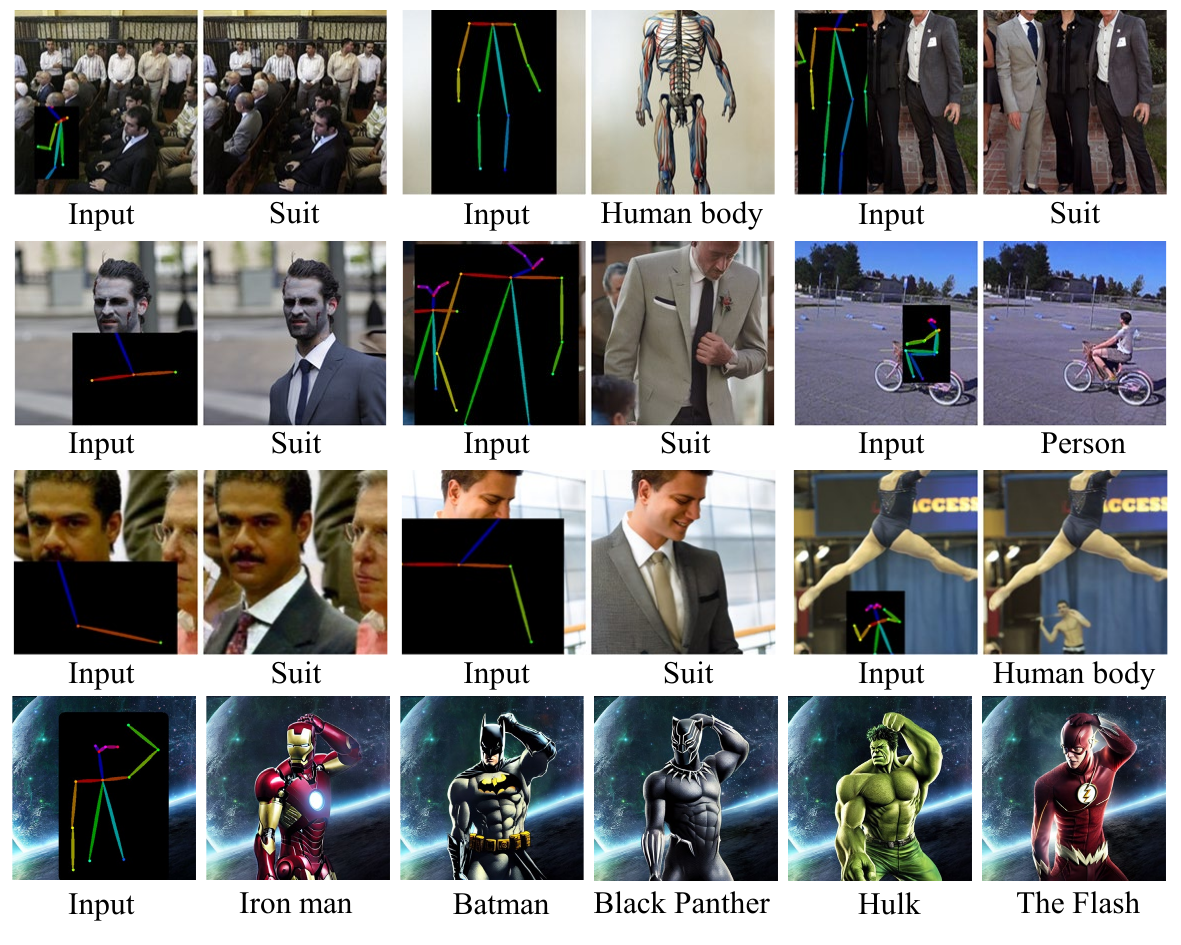}
    \caption{Visual results of \modelname\ with the ControlNet conditioned on poses.}
    \label{fig:sup_pose}
\end{figure*}

We evaluated the compatibility of \modelname\ with various ControlNets \cite{zhang2023adding}, enabling users to incorporate additional conditions for guiding the inpainting process. We tested four ControlNets: canny edge\footnote{https://huggingface.co/lllyasviel/sd-controlnet-canny}, depth\footnote{https://huggingface.co/lllyasviel/sd-controlnet-depth}, hed boundary\footnote{https://huggingface.co/lllyasviel/sd-controlnet-hed}, and human pose\footnote{https://huggingface.co/lllyasviel/sd-controlnet-openpose}. Our results, shown in Figures \ref{fig:sup_canny} to \ref{fig:sup_pose}, demonstrate that \modelname\ effectively generates high-quality images aligned with the provided ControlNet conditions. This highlights the versatility of \modelname\ in leveraging existing ControlNets for achieving controllable image inpainting.


%
%
\bibliographystyle{splncs04}
\bibliography{11_references}
\end{document}